# Image-Based Multi-UAV Tracking System in a Cluttered Environment

Hsin-Ai Hung, Hao-Huan Hsu, and Teng-Hu Cheng

***Abstract*—A tracking controller for unmanned aerial vehicles (UAVs) is developed to track moving targets undergoing unknown translational and rotational motions. The main challenges are to control both the relative positions and angles between the target and the UAVs to within desired values, and to guarantee that the generated control inputs to the UAVs are feasible (i.e., within their motion capabilities). Moreover, the UAVs are controlled to ensure that the target always remains within the fields of view of their onboard cameras. To the best of our knowledge, this is the first work to apply multiple UAVs to cooperatively track a dynamic target while ensuring that the UAVs remain connected and that both occlusion and collisions are avoided. To achieve these control objectives, a designed controller solved based on the aforementioned tracking controller using quadratic programming can generate minimally invasive control actions to achieve occlusion avoidance and collision avoidance. Furthermore, control barrier functions (CBFs) with a distributed design are developed in order to reduce the amount of inter-UAV communication. Simulations were performed to assess the efficacy and performance of the developed CBF-based controller for the multi-UAV system in tracking a target.**

***Index Terms*—Unscented Kalman filter, Estimation, Tracking of moving targets, UAV**

## I. Introduction

The agile motion capabilities of unmanned aerial vehicles (UAVs) are being utilized to perform an increasingly diverse range of tasks, such as surveillance [1], [2], aerial photography [3], [4], and rescue [5]. One of the critical issues in completing these tasks is target tracking. Previous work [6] has demonstrated that a UAV is capable of tracking a dynamic and temporarily-occluded target undergoing unknown motions. However, tracking failure can occur when the occlusion period is too long to accurately predict the target motion. Using a single UAV to track a target can also lead to mission failure due to loss of tracking caused by occlusion. In contrast, combining multiple UAVs and inter-UAV communication can increase the robustness of tracking tasks in many applications, such as tracking a moving target from different angles [7] and tracking multiple moving targets [8].

Visual servo control has been developed for tracking objects, where knowledge of the target motion is important for executing tracking control using UAVs. However, in many cases the motion of the target is unknown and difficult to predict. Many works have addressed this issue using visual errors for feedback control to keep the target within the field of view (FOV) [9]. For example, [10] and [11] determined the position of a target in an image frame based on its specific color and shape. Bounding boxes derived from an image tracking system (e.g., KCF-TLD) or a deep neural network (DNN) have also been utilized as a feature for feedback control [6], [12]–[14]. However, these previous works focused on either object detection or visual feedback control, and did not consider collision avoidance and occlusion avoidance.

It is crucial to avoid obstacle collision and occlusion in a cluttered environment during a mission.A collision can cause tracking failure and occlusion can cause loss of the target image that leads to failure of state estimation of the target undergoing unknown motion. In [15]–[17], collision avoidance was addressed by including a term in the cost function that considered distances from an obstacle. In [3] and [4], occlusion avoidance was addressed by solving a nonconvex optimization problem. However, collisions and occlusion may still occur since they were only considered as soft constraints [17]. This issue can be solved by designing control barrier functions (CBFs) [18], [19] as hard constraints in the optimization problem, which guarantees system safety as long as the control inputs and the optimization problem can be solved under the safety constraint. Additionally, the computational cost is important when designing controllers with the capability of obstacle avoidance, and this can be reduced by leveraging different mathematical characteristics. For example, a minimally invasive collision-avoidance controller was introduced in [20], where the controller can change the original control program only when it is absolutely necessary. However, the heavy computation load associated with solving the Hamilton-Jacobi-Bellman equations limits the applicability of that approach to large-scale multirobot systems.

CBFs are similar to control Lyapunov functions in the sense that they guarantee the forward invariance of the safety set without computing the system's reachable set. The use of CBFs for coordinating multirobot systems has been studied in recent years [18], [19]. CBFs were first applied to guarantee system safety in applications such as adaptive cruise control [21], legged robots [22], and the Segway vehicle [23]. CBFs have also been utilized to maintain the connectivity of a multirobot network [24], [25]. Multiple control objectives for robots were also addressed, such as

Department of Mechanical Engineering, National Yang Ming Chiao Tung University, Hsinchu 30010, Taiwan. Email: louise.me08g@nctu.edu.tw, water4202@gmail.com, tenghu@g2.nctu.edu.tw

This research is supported by the Ministry of Science and Technology, Taiwan (Grant Number MOST 110-2222-E-A49 -005 -) and by Pervasive Artificial Intelligence Research (PAIR) Labs, Taiwan (Grant Number MOST 110-2634-F-009 -018 -).

collision avoidance and connectivity maintenance [26] , [27]. To increase the fault tolerance and guarantee the safety of robots based on information from neighboring robots, distributed CBFs were developed in [28], [29]. Motivated by the aforementioned works, distributed CBFs are designed in this work for multiple UAVs to cooperatively track a moving target while ensuring collision avoidance, connectivity maintenance, and occlusion avoidance. A target tracking controller is first developed based on image feedback, where You Only Look Once (YOLO) DNN is designed to generate a bounding box and determine the relative angle of the target as features for tracking control. To further meet the constraints of collision avoidance, occlusion avoidance, and connectivity maintenance, a modified controller designed based on a tracking controller using quadratic programming (QP) is developed to generate minimally invasive control actions to ensure occlusion avoidance and collision avoidance. The main contributions of this work can be summarized as follows:

1) In contrast to previous approaches, the relative rotation between the target and the UAV is modeled and can be controlled to a desired value.
2) Multiple UAVs are employed to cooperatively track a dynamic target by communicating its position among neighboring UAVs for feedback to ensure collision avoidance and connectivity maintenance.
3) A multiobjective optimization problem based on CBFs is developed to ensure not only collision avoidance and connectivity maintenance, but also avoiding occlusion by obstacles.
4) The velocity of the UAV not only restricted by the CBF, but also determined by the dynamics of the desired relative angle. Therefore, the interplay between the estimation, control, and constraints are determined by the developed optimal problem.
5) There are many ways to design the CBF functions, but to ensure that the controller is distributed, a sufficient condition needs to be developed and analyzed. For example, the condition $\gamma_s$ defined in (28) is designed to ensure the designed controller is distributed.
6) Moreover, the CBFs and the controllers are co-designed to ensure the existence of the barrier functions and to ensure that the forward invariance property is preserved based on [18]. Therefore, as long as the constraints can be satisfied, the optimal solution can be solved from the cost function.
7) Slack variables are also added and used to determine if the constraints are too conservative.

The report is organized as follows: Section II formulates the multi-UAV tracking problem and describes the model for kinematics and dynamics between the UAVs and the target. Section III designs nonlinear model predictive control (NMPC) for target tracking. Section IV introduces CBFs for guaranteeing the safety of the multi-UAV system. In Section V, an optimization problem based on the CBFs as constraints is constructed and solved for the minimally invasive controller. The simulation results are presented in Section VI, and conclusion are drawn and future work is summarized in Section VII.

## II. Preliminaries

### A. Motivation

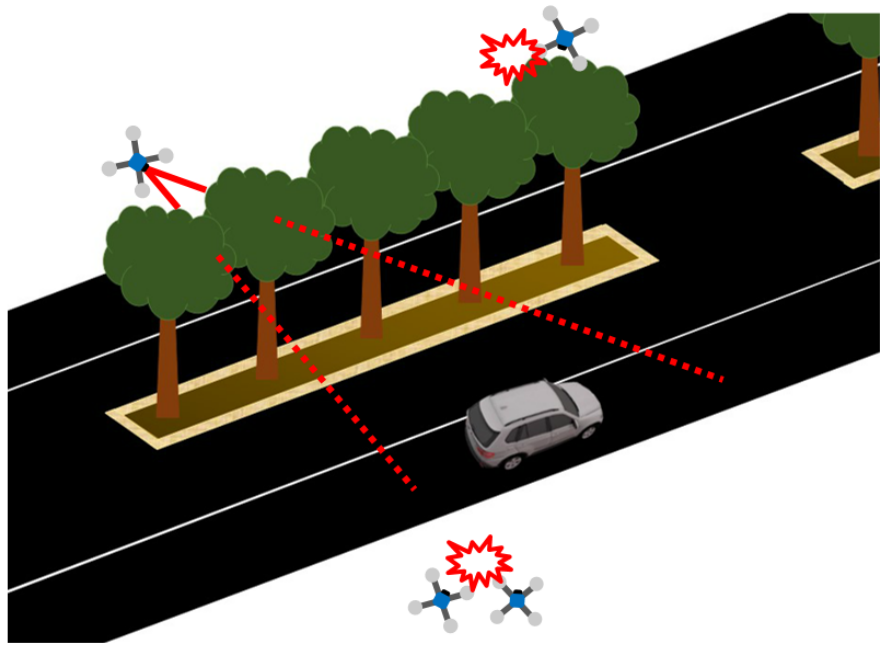

Fig. 1. Multiple UAVs tracking a moving target in a cluttered environment.

Fig. 1 illustrates the difficulties encountered when multiple UAVs are tracking a moving target in a cluttered environment. The tracking mission will fail if the UAVs collide with each other or with the static obstacles (e.g., trees). Moreover, occlusion of the target by obstacles will prevent the UAVs from detecting the target, and leading to failure of the tracking mission. To solve these problems, CBFs for (i) collision avoidance (ii) connectivity maintenance, and (iii) occlusion avoidance are introduced in this work.

### B. Kinematics Model

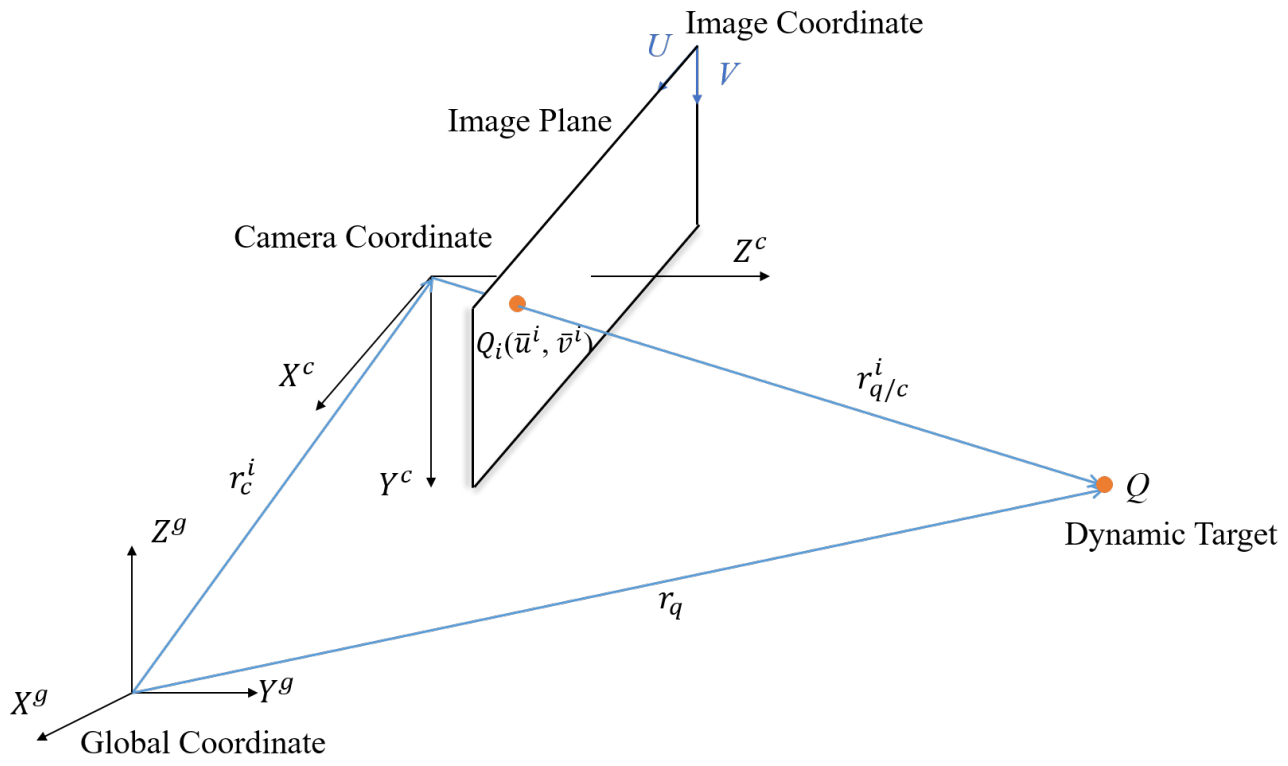

Fig. 2. Model for the kinematics between the camera and the target.

Consider a team of $n$ UAVs in the system as depicted in Fig. 2, which shows the relationship between a moving target and an onboard camera fixed on the $i^{\text{th}}$ UAV, where $i \in N \triangleq \{1,\ 2,\ \ldots,\ n\}$. $r^i_q = \begin{bmatrix} x^i_q & y^i_q & z^i_q \end{bmatrix}^{\text{T}}$ denotes the position of the moving target estimated from the $i^{\text{th}}$ UAV, and $r^i_c = \begin{bmatrix} x^i_c & y^i_c & z^i_c \end{bmatrix}^{\text{T}}$ denotes the position of the camera fixed on the $i^{\text{th}}$ UAV, which can be measured by the embedded

GPS or motion-capture systems, all expressed in the camera frame. The relative position between the moving target and the camera is defined as

$$r^i_{q/c} = r^i_q - r^i_c, \tag{1}$$

where $r^i_{q/c} = \begin{bmatrix} X^i & Y^i & Z^i \end{bmatrix}$ is defined in the camera frame. Taking the time derivative on both sides of (1) yields the relative velocity as

$$\dot{r}^i_{q/c} \;=\; V^i_q - V^i_c - \omega^i_c \times r^i_{q/c}, \tag{2}$$

where $V^i_q = \begin{bmatrix} v^i_{qx} & v^i_{qy} & v^i_{qz} \end{bmatrix}^{\mathrm{T}}$ is the translational velocity of the dynamic target, and $V^i_c \triangleq \begin{bmatrix} v^i_{cx} & v^i_{cy} & v^i_{cz} \end{bmatrix}^{\mathrm{T}}$ and $\omega^i_c \triangleq \begin{bmatrix} \omega^i_{cx} & \omega^i_{cy} & \omega^i_{cz} \end{bmatrix}^{\mathrm{T}}$ are the translational and angular velocities of the camera, respectively, both of which represent the control commands to be designed. Each of the UAVs is located at $p^i = (r^i_c)^g \in \mathbb{R}^3$ with single-integrator dynamics

$$\dot{p}^i = (V^i_c)^g \in \mathbb{R}^3, \tag{3}$$

where the $(\cdot)^g$ superscript denotes the vector defined in the global frame, and the stacked position of the team of UAVs is denoted as $p = \left[p^1,\ \ldots,\ p^n\right] \in \mathbb{R}^{3n}$, along with control command $(u^i)^g = [(V^i_c)^g\ (\omega^i_c)^g]^T \in \mathbb{R}^6$.

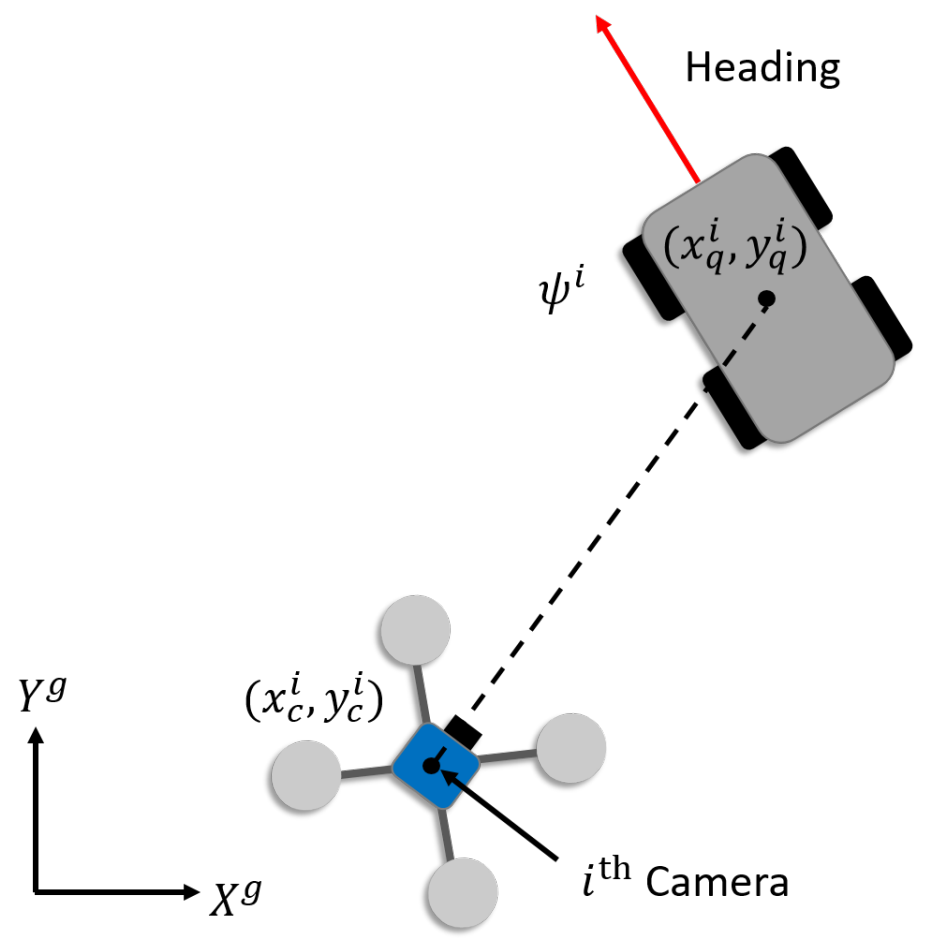

Fig. 3. $\psi^i$ is defined as the angle between the heading of the target and the relative position from the target to the $i^{th}$ UAV in the global frame.

Tracking a target undergoing 2D motion at a specific relative location requires knowledge of the relative angle of the target denoted as $\psi^i \in \mathbb{R}$ (Fig. 3):

$$\psi^i = \arctan\left(\frac{(y^i_c)^g - (y^i_q)^g}{(x^i_c)^g - (x^i_q)^g}\right) - \arctan\left(\frac{(v^i_{qy})^g}{(v^i_{qx})^g}\right). \tag{4}$$

**Assumption 1.** The trajectory of the target is unknown but bounded; that is, $r_q, \dot{r}_q, \ddot{r}_q \in \mathcal{L}_\infty$.

To estimate the state of the target, a dynamics model and a measurement model are introduced in Section II-C and II-D.

### C. Dynamics Model

Since the dynamics of the camera and the target are coupled, the states of the overall system are defined as

$$\mathbf{x}^i = \left[x^i_1,\ x^i_2,\ x^i_3,\ \psi^i,\ (r^i_q)^T,\ (V^i_q)^T\right]^{\mathrm{T}}, \tag{5}$$

where

$$\left[x^i_1,\ x^i_2,\ x^i_3\right] = \left[\frac{X^i}{Z^i},\ \frac{Y^i}{Z^i},\ \frac{1}{Z^i}\right] \tag{6}$$

are the states defined to facilitate the subsequent analysis. The dynamics of $\mathbf{x}^i$ can be obtained by taking the time derivative on both sides of (5):

$$\dot{\mathbf{x}}^i = \begin{bmatrix} v^i_{qx}x^i_3 - v^i_{qz}x^i_1x^i_3 + \zeta^i_1 + \eta^i_1 \\ v^i_{qy}x^i_3 - v^i_{qz}x^i_2x^i_3 + \zeta^i_2 + \eta^i_2 \\ -v^i_{qz}(x^i_3)^2 + v^i_{cz}(x^i_3)^2 - (\omega^i_{cy}x^i_1 - \omega^i_{cx}x^i_2)x^i_3 \\ \dot{\psi}^i \\ V^i_q \\ 0_{3\times1} \end{bmatrix}, \tag{7}$$

where (2) is used and $\zeta^i_1,\ \zeta^i_2,\ \eta^i_1,\ \eta^i_2 \in \mathbb{R}$ are defined as

$$\begin{aligned} \zeta^i_1 &= \omega^i_{cz}x^i_2 - \omega^i_{cy} - \omega^i_{cy}\left(x^i_1\right)^2 + \omega^i_{cx}x^i_1x^i_2 \\ \zeta^i_2 &= -\omega^i_{cz}x^i_1 + \omega^i_{cx} + \omega^i_{cx}\left(x^i_2\right)^2 - \omega^i_{cy}x^i_1x^i_2 \\ \eta^i_1 &= (v^i_{cz}x^i_1 - v^i_{cx})x^i_3 \\ \eta^i_2 &= (v^i_{cz}x^i_2 - v^i_{cy})x^i_3. \end{aligned}$$

### D. Measurement Model

This section defines two measurements performed using a YOLO DNN. The bounding box and the relative angle measurement provide information to correct the states in the unscented Kalman filter (UKF) developed in Section II-E. A detailed description of the DNN can be found in [6].

*a) Bounding Box and Relative Angle:* Based on the pinhole model, states $x^i_1$ and $x^i_2$ can be obtained by projecting the position of the target onto the image plane as

$$\begin{aligned} x^i_1 &= \frac{\bar{u}^i - c_u}{f_x} \\ x^i_2 &= \frac{\bar{v}^i - c_v}{f_y} \\ x^i_3 &= \frac{1}{d^i}, \end{aligned} \tag{8}$$

where $f_x$ and $f_y$ are the focal lengths, $c_u$ and $c_v$ represent the center of the image, $\bar{u}^i$ and $\bar{v}^i$ represent the center of the detected bounding box of the moving target along the $U, V$ direction in the image frame (see Fig. 2), and $d^i$ is the depth between the target and the $i^{\text{th}}$ UAV. Therefore, based on (1), (6), and (8), the measurement model can be derived as

$$h(\mathbf{x}^i) = \begin{bmatrix} f_x x^i_1 + c_u \\ f_y x^i_2 + c_v \\ 1/x^i_3 \\ \psi^i \\ r^i_c \end{bmatrix}, \tag{9}$$

where $\psi^i$ is also measured by the YOLO as shown in Fig. 3.

*Remark* 1. Changes in the target velocity can result in the developed UKF making inaccurate state estimations. By considering the model mismatch between the real and constant-velocity models, defined in (7), as process noise in the UKF, the states can be estimated and updated to approach the ground truth.

### E. Unscented Kalman Filter

A UKF [30] is applied in this section to estimate the state of dynamic systems with noisy or intermittent measurements, since a UKF can outperform an extended Kalman filter in several ways: dealing with highly nonlinear systems, accurate to two terms of the Taylor expansion, and higher efficiency. By discretizing (7) and (9) using the forward Euler method and including noise, the nonlinear dynamic system can be expressed as

$$\mathbf{x}_{k+1}^i = F(\mathbf{x}_k^i, \mathbf{u}_k^i) + w_k \tag{10}$$

$$\mathbf{z}_k^i = H(\mathbf{x}_k^i) + v_k, \tag{11}$$

where $w_k$ and $v_k$ represent the process and measurement noise, respectively, $\mathbf{x}_k^i$ is the discretized states of the overall system defined in (5), $\mathbf{u}_k^i$ is the discretized control input, $F(\cdot)$ is the dynamics model used for prediction, $H(\cdot)$ is the measurement model used for updating, and $\mathbf{z}_k^i$ is the discretized measurement defined as

$$\mathbf{z}_k^i = \begin{bmatrix} \bar{u}_z^i \\ \bar{v}_z^i \\ d_z^i \\ \psi_z^i \\ r_{c,z}^i \end{bmatrix},$$

where the $z$ subscript represents the measured value of the denoted argument. When the bounding box cannot be detected, the state is predicted by the dynamics model using (10), which is reliable over a short period of time before the detection process recovers.

*Remark* 2. The UKF-based method developed in Section II-E utilizes a high-order polynomial to continuously estimate the unknown motion of a moving target; that is, the polynomial is found based on a QP method to fit the position and velocity data over the latest time period (cf. (27) of [6]).

## III. Nonlinear Model Predictive Control

An NMPC is designed in this section and is then utilized in Section V to compute the minimally invasive controller. The control input of the NMPC is calculated by optimizing the cost function designed in this section that considers not only the current and upcoming feature errors, but also the control input. Using the NMPC can decrease both the aggressiveness of the flying behavior and the energy consumption.

### A. Transformation of the Dynamics Model

Since computing the dynamics defined in (7) is too inefficient, the rotational dynamics is transformed to increase the computational efficiency. On the other hand, to avoid the UAVs reaching their maximum speed during tracking, slow time variations on the angular velocity of the target are also considered in Assumption 2.

**Assumption 2.** The target is initially located within the FOV of the camera, and the angular velocity of the target exhibits only slow time variations (i.e., $\omega_q \approx 0$). During the steady state, the UAV is controlled such that its camera is looking almost directly at the target (i.e., $V_q \approx V_c$), as shown in Fig. 4. In other words, the heading of the camera is parallel to the direction of the position vector from the camera to the target.

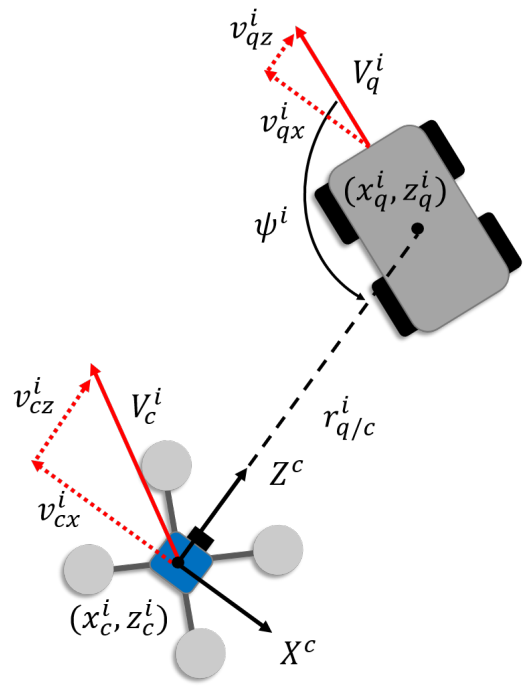

Fig. 4. Directions of velocities expressed in the camera frame.

Based on Assumption 2, velocities $V_q^i$ and $V_c^i$ defined in (2) satisfy $V_q^i \approx V_c^i$ and $\omega_q^i \approx 0$, which implies $\dot{\psi}^i \approx \left(\omega_{cz}^i\right)^g = -\omega_{cy}^i$. Therefore, the dynamics of relative angle $\dot{\psi}^i$ can be expressed as

$$\dot{\psi}^i = -\omega_{cy}^i = \frac{-(v_{qx}^i - v_{cx}^i)}{\left\| r_{q/c}^i \right\|},$$

which is equivalent to

$$\dot{\psi}^i = \frac{(v_{cx}^i - v_{qx}^i)x_3^i}{\sqrt{1 + (x_1^i)^2 + (x_2^i)^2}} \tag{12}$$

based on the relation

$$\begin{aligned} \left\| r_{q/c}^i \right\| &= \sqrt{(X^i)^2 + (Y^i)^2 + (Z^i)^2} \\ &= \frac{\sqrt{1 + (x_1^i)^2 + (x_2^i)^2}}{x_3^i}. \end{aligned} \tag{13}$$

The dynamics model of the overall system defined in (7) can

be replaced with

$$\dot{\mathbf{x}}^i = \begin{bmatrix} v_{qx}^i x_3^i - v_{qz}^i x_1^i x_3^i + \zeta_1 + \eta_1 \\ v_{qy}^i x_3^i - v_{qz}^i x_2^i x_3^i + \zeta_2 + \eta_2 \\ -v_{qz}^i (x_3^i)^2 + v_{cz}^i (x_3^i)^2 - (\omega_{cy}^i x_1^i - \omega_{cx}^i x_2^i) x_3^i \\ \frac{(v_{cx}^i - v_{qx}^i) x_3^i}{\sqrt{1+(x_1^i)^2+(x_2^i)^2}} \\ V_q^i \\ 0_{3\times 1} \end{bmatrix}. \tag{14}$$

*Remark* 3. Assumption 2 might not always be satisfied, in which cases the mismatches can be considered as process noise and be corrected using the relative angle as measured by the YOLO network.

### B. Nonlinear Model Predictive Control

Only part of the dynamics model in (14) is required to further design the controller; that is, given the UAV constraints $\omega_{cx}^i \approx 0$ and $\omega_{cz}^i \approx 0$ according to [10], the control input can be reduced to $u^i \triangleq \left[v_{cx}^i,\, v_{cy}^i,\, v_{cz}^i,\, \omega_{cy}^i\right]^T \in \mathbb{R}^4$, and the dynamics model of feature vector $s_m^i = \left[x_1^i,\, x_2^i,\, x_3^i,\, \psi^i\right]^{\mathrm{T}} \in \mathbb{R}^4$ used for prediction in the NMPC can be written as

$$\dot{s}_m^i = \begin{bmatrix} -x_3^i (v_{cx}^i - v_{qx}^i) + x_1^i x_3^i (v_{cz}^i - v_{qz}^i) - (1 + (x_1^i)^2) \omega_{cy}^i \\ -x_3^i (v_{cy}^i - v_{qy}^i) + x_2^i x_3^i (v_{cz}^i - v_{qz}^i) - \omega_{cy}^i x_1^i x_2^i \\ (x_3^i)^2 (v_{cz}^i - v_{qz}^i) - \omega_{cy}^i x_1^i x_3^i \\ \frac{(v_{cx}^i - v_{qx}^i) x_3^i}{\sqrt{1+(x_1^i)^2+(x_2^i)^2}} \end{bmatrix}. \tag{15}$$

To deal with the modeling errors associated with using the transformed dynamics model and the imperfect actuation of the $i^{\text{th}}$ UAV, an error signal $\epsilon_k$ is designed as

$$\epsilon_k \;=\; s_k^i - s_{m,k}^i, \tag{16}$$

where feature vector $s_k^i$ is estimated using the UKF, and feature vector $s_{m,k}^i$ is predicted using the transformed model (15) at time $t_k$. The desired feature vector utilized in the NMPC can be defined as

$$s_{d,k}^i \;=\; s^{i,*} - \epsilon_k, \tag{17}$$

where $s^{i,*} \triangleq [x_1^*,\, x_2^*,\, x_3^*,\, \psi^*]^{\mathrm{T}} \in \mathbb{R}^4$ is the reference feature vector prescribed by the user.

With (15) and (17), the optimal control problem (OCP) of the NMPC at time $t_k$ can be expressed as

$$\min_{s_m^i, u^i} \sum_{n=0}^{N_p-1} \left[ \left\| s_{m,k+n}^i - s_{d,k+n}^i \right\|_{Q_s}^2 + \left\| u_{k+n}^i - u_{d,k+n}^i \right\|_{R_u}^2 \right] + \left\| s_{m,k+N_p}^i - s_{d,k+N_p}^i \right\|_{W_s}^2, \tag{18}$$

s.t.

$$\begin{aligned} & s_{m,k+n+1}^i = F_m(s_{m,k+n}^i,\, u_{k+n}^i),\; n = 0, \ldots, N_p - 1 \\ & u_{d,k+n+1}^i = u_{d,k+n}^i + \begin{bmatrix} \ddot{\hat{\Gamma}}_k^i, & 0 \end{bmatrix}^T \mathrm{d}t,\; n = 0, \ldots, N_p - 1 \\ & s_{m,k+n}^i \in \mathbb{S},\; n = 1, \ldots, N_p \\ & u_{k+n}^i \in \mathbb{U},\; n = 0, \ldots, N_p - 1 \\ & s_{d,k+n}^i = s^* - \epsilon_k,\; n = 0, \ldots, N_p \\ & s_{m,k}^i = s_k^i, \end{aligned} \tag{19}$$

where $u_d^i \in \mathbb{R}^4$ is defined as the desired velocity that control input $u^i$ needs to achieve, and can be updated by $\ddot{\hat{\Gamma}}_k^i$ (as defined in Section 5 of [6]) over the prediction horizon, as depicted in Fig. 5. $S_m^i = \{s_{m,k}^i,\, s_{m,k+1}^i,\, \ldots,\, s_{m,k+N_p}^i\}$ denotes the predicted features at sampling time $t_k$ based on control sequence $U^i = \left\{u_k^i,\, u_{k+1}^i,\, \ldots,\, u_{k+N_p-1}^i\right\}$ and estimation $s_k^i$ obtained using the UKF. $Q_s$, $R_u$, and $W_s$ defined in (18) are positive weighting matrices, and $N_p \in \mathbb{N}$ is the prediction horizon. Increasing $N_p$ decreases the aggressiveness of the control inputs but also increases the computational effort. $F_m(\cdot)$ is the model defined in (15), and $\mathbb{S}$ and $\mathbb{U}$ are the sets of constraints of the feature vector and the control input given by

$$\mathbb{S} \;\triangleq\; \left\{ s^i \in \mathbb{R}^4 \middle| \begin{bmatrix} x_{1,\min} \\ x_{2,\min} \\ x_{3,\min} \\ \psi_{\min} \end{bmatrix} \le s^i \le \begin{bmatrix} x_{1,\max} \\ x_{2,\max} \\ x_{3,\max} \\ \psi_{\max} \end{bmatrix} \right\}$$

$$\mathbb{U} \;\triangleq\; \left\{ u^i \in \mathbb{R}^4 \middle| \begin{bmatrix} v_{cx,\min} \\ v_{cy,\min} \\ v_{cz,\min} \\ \omega_{cy,\min} \end{bmatrix} \le u^i \le \begin{bmatrix} v_{cx,\max} \\ v_{cy,\max} \\ v_{cz,\max} \\ \omega_{cy,\max} \end{bmatrix} \right\}.$$

Error signal $\epsilon_k$ defined in (16) is assumed to be constant over the prediction horizon, and $s_{d,k+n}^i$, $n = 0, \ldots, N_p$, would be constant over the prediction horizon.

The direct multiple shooting method is employed to solve the problem since lifting the OCP to a higher dimension can usually speed up the rate of convergence. The OCP is repeated at every sampling time, with only the first vector $u_k^i$ in control set $U^i$ being adopted in the system. Several software libraries can be used to solve this kind of nonlinear programming problem (NLP); the CasADi tool is used in this work for formulating the NLP, and the Ipopt library is used to solve it.

*Remark* 4. A control architecture was developed for general target tracking without limitation to vehicles. Therefore, given a bounding box enclosing the target in the image, measurements of the feature vector and relative angle (i.e., $s^i$) can be used to predict the motion of target $v_q^i$, which can improve the tracking performance and, in turn, enhance the detection of bounding boxes.

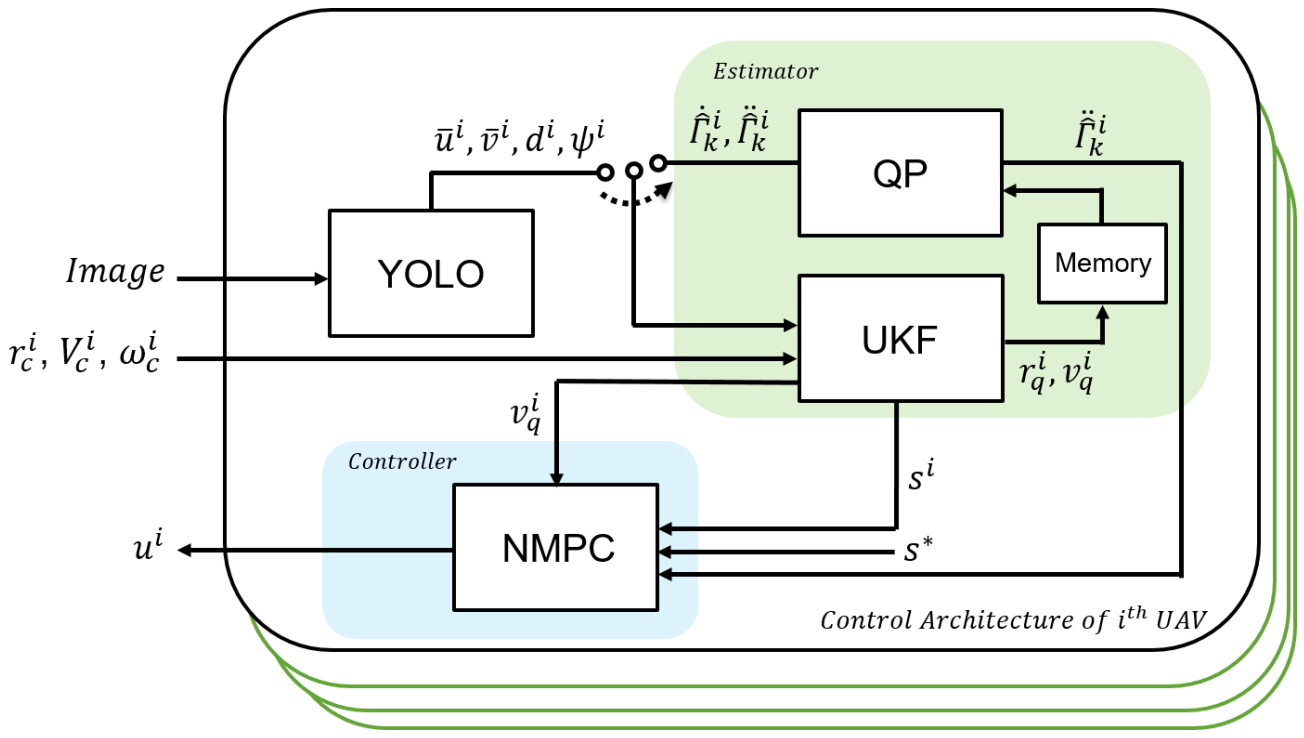

Fig. 5. Block diagram of the tracking controller for the $i^{\text{th}}$ UAV. A switch is added to select one of the inputs to the UKF depending on the presence of the target in the image. When the measurement data $[\bar{u}_z^i,\ \bar{v}_z^i,\ d_z^i,\ \psi_z^i]$ are not available due to occlusion, the motion obtained by QP (i.e., $\hat{\dot{\Gamma}}_k^i$ and $\hat{\ddot{\Gamma}}_k^i$) is used to predict the target motion for the UKF. Otherwise, the measured values obtained from the detected bounding box $[\bar{u}_z^i,\ \bar{v}_z^i,\ d_z^i,\ \psi_z^i]$ are imported into the UKF for updating.

While using a decentralized NMPC controller $u^i$ ensures the local optimal configuration of the $i^{\text{th}}$ UAV, this approach might still lead to mission failure due to UAVs colliding or a disconnected network, since the UAVs need to communicate their positions for their controller to achieve collision avoidance. Therefore, CBFs are introduced to guarantee collision avoidance, occlusion avoidance, and connectivity maintenance when computing control command $u^i$ developed in (18).

## IV. Control Barrier Functions

The CBF has been introduced as a tool for guaranteeing the safety of a system [21]. A typical CBF is designed to be a forward invariant set, so that the solved control command can always keep the closed-loop system in the safety set if it starts in that set [26]. To this end, CBFs for collision avoidance, connectivity maintenance, and occlusion avoidance are developed in this section.

### A. CBF for Collision Avoidance

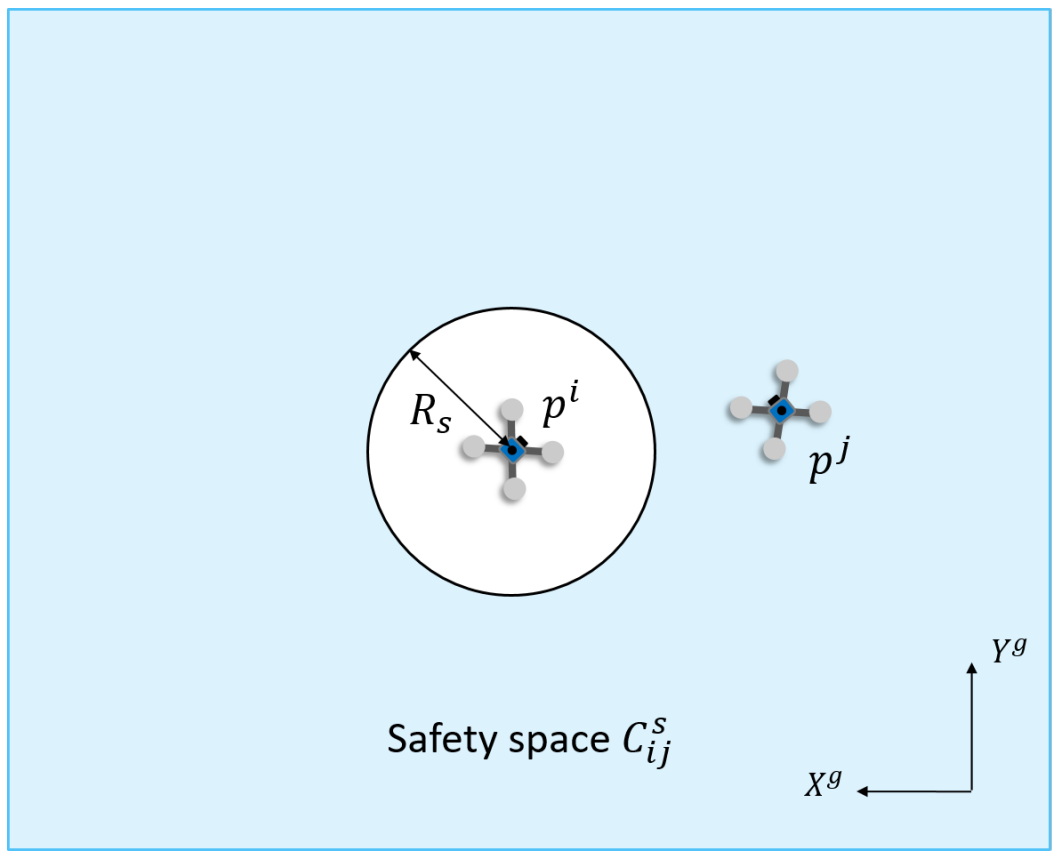

Fig. 6. $R_s$ denotes the minimum safety range, and the blue zone is safety space $C_{ij}^s$ where a potential collision between the $i^{\text{th}}$ and $j^{\text{th}}$ UAVs can be avoided.

To avoid collisions between UAVs, a safety space denoted as $C_{ij}^s,\ \forall i,j \in N,\ i \neq j$ is defined as

$$C_{ij}^s = \left\{p \in \mathbb{R}^{3n} \mid h_{ij}^s(p) = \| p^i - p^j \|^2 - R_s^2 \geq 0\right\}, \quad (20)$$

where $C_{ij}^s$ is the set as shown in Fig. 6. Function $h_{ij}^s$ is regarded as a CBF that satisfies

$$\dot{h}_{ij}^s(p) \geq -\gamma_s h_{ij}^s, \quad (21)$$

where $\gamma_s$ is a user-defined positive constant. With (20) and the single-integrator dynamics defined in (3), (21) can be further expressed as

$$-2(p^i - p^j)(V_c^i)^g + 2(p^i - p^j)(V_c^j)^g \leq \gamma_s h_{ij}^s, \quad (22)$$

which can be separated into the following sufficient conditions [29]:

$$-2(p^i - p^j)(V_c^i)^g \leq \frac{1}{2}\gamma_s h_{ij}^s, \quad (23)$$

$$2(p^i - p^j)(V_c^j)^g \leq \frac{1}{2}\gamma_s h_{ij}^s. \quad (24)$$

Inequalities (23) and (24) can be transformed into the following linear inequality:

$$A_{ij}^s(u^i)^g \leq b_{ij}^s,\ \forall i,j \in N,\ i \neq j,$$

where

$$A_{ij}^s = \left[\underbrace{-2(p^i - p^j)^{\mathrm{T}},\ 0,\ 0,\ 0}_{i^{\text{th}}\text{UAV}}\right], \qquad b_{ij}^s = \frac{1}{2}\gamma_s h_{ij}^s. \quad (25)$$

In other words, the $i^{\text{th}}$ and $j^{\text{th}}$ UAVs can always avoid colliding if they are initially in safety space $C_{ij}^s$ and control input $(u^i)^g$ lies within admissible control set $K_{ij}^{sN}$ defined as

$$K_{ij}^{sN} = \left\{(u^i)^g \in \mathbb{R}^6 \mid A_{ij}^s(u^i)^g \leq b_{ij}^s,\ \forall i,j \in N,\ i \neq j\right\}, \quad (26)$$

where $K_{ij}^{sN}$ is the admissible control space obtained by the mapping of the safety space defined in (20). It is shown in [31] that if control inputs $(u^i)^g$ stay in $K_{ij}^{sN}(p)$, the UAVs stay within safety space $C_{ij}^s$ all of the time, and hence the forward invariance property of $C_{ij}^s$ is guaranteed.

*Remark* 5. Admissible control set $K_{ij}^{sN}$ defined in (26) ensures the safety of the entire UAV team; however, it requires position information from all of the UAVs (i.e., $j \in N$) in the network, and the amount of communication required to achieve this will increase with the number of UAVs. This makes it necessary to design a decentralized CBF based only on information about neighboring UAVs.

To this end, (26) is modified as

$$K_{ij}^{s} = \left\{ (u^{i})^{g} \in \mathbb{R}^{6} \mid A_{ij}^{s}(u^{i})^{g} \leq b_{ij}^{s},\ \forall i \in N,\ j \in N_{si} \right\}, \tag{27}$$

$$\gamma_s \geq \frac{4\alpha_v R_s}{D_s^2 - R_s^2}, \tag{28}$$

where $N_{si} = \left\{ j \in N \mid \| p^i - p^j \| \leq D_s,\ \forall i \neq j \right\}$ is the set of the $i^{\text{th}}$ UAV's neighbors that collisions start to be considered, and $D_s$ is the distance, which satisfies $R_s < D_s \leq R_c$, where $R_c \in \mathbb{R}$ is the maximum connectivity range. If $\gamma_s$ satisfies (28), it is proven in [32] that safety is still guaranteed by admissible control set $K_{ij}^{s}$ defined in (27), where $\alpha_v$ is the upper bound of the translational velocities of the camera $\| (V_c^i)^g \| \leq \alpha_v,\ \forall i \in N$. The CBF can also be utilized to avoid collisions between the static obstacles and the UAVs.

### *B. CBF for Connectivity Maintenance*

In order to ensure inter-UAV communications, the distance between the $i^{\text{th}}$ and $j^{\text{th}}$ UAVs needs to be maintained to within maximum connectivity range $R_c$, which is defined by connectivity space $C_{ij}^{c},\ \forall i \in N,\ j \in N_{ci}$ as

$$C_{ij}^{c} = \left\{ p \in \mathbb{R}^{3n} \mid h_{ij}^{c}(p) = R_c^2 - \| p^i - p^j \|^2 \geq 0 \right\}, \tag{29}$$

and the neighbor set of the $i^{\text{th}}$ UAV is defined as

$$N_{ci} = \{ j \in N \mid \| p^i - p^j \| \leq R_c, \forall i \neq j \}.$$

It is assumed that the $i^{\text{th}}$ UAV can obtain the positions of its neighboring UAVs once they are inside connectivity range $R_c$. Function $h_{ij}^{c}$ is considered to be a CBF that satisfies

$$\dot{h}_{ij}^{c}(p) \geq -\gamma_c h_{ij}^{c}, \tag{30}$$

where $\gamma_c$ is a user-defined parameter for fine tuning the connectivity space defined in (29). With (29) and the single-integrator dynamics defined in (3), (30) needs to satisfy

$$2(p^i - p^j)(V_c^i)^g - 2(p^i - p^j)(V_c^j)^g \leq \gamma_c h_{ij}^{c}, \tag{31}$$

which can be separated into the following sufficient conditions [29]:

$$2(p^i - p^j)(V_c^i)^g \leq \frac{1}{2}\gamma_c h_{ij}^{s}, \tag{32}$$

$$-2(p^i - p^j)(V_c^j)^g \leq \frac{1}{2}\gamma_c h_{ij}^{s}. \tag{33}$$

Inequalities (32) and (33) can be further transformed into the following linear inequalities:

$$A_{ij}^{c}(u^i)^g \leq b_{ij}^{c},\ \forall i \in N,\ j \in N_{ci},$$

where $A_{ij}^{c}$ and $b_{ij}^{c}$ are defined as

$$\begin{aligned} A_{ij}^{c} &= \Bigg[ \underbrace{2(p^i - p^j)^{\mathrm{T}},\ 0,\ 0,\ 0}_{i^{\text{th}}\text{UAV}} \Bigg], \\ b_{ij}^{c} &= \frac{1}{2}\gamma_c h_{ij}^{c}. \end{aligned} \tag{34}$$

Based on (34), the $i^{\text{th}}$ and $j^{\text{th}}$ UAVs will always stay connected if they are initially connected, and control input $(u^i)^g$ inside admissible control space $K_{ij}^{c}$ is defined as

$$K_{ij}^{c} = \left\{ (u^i)^g \in \mathbb{R}^6 \mid A_{ij}^{c}(u^i)^g \leq b_{ij}^{c},\ \forall i \in N,\ j \in N_{ci} \right\}. \tag{35}$$

### *C. CBF for Occlusion Avoidance*

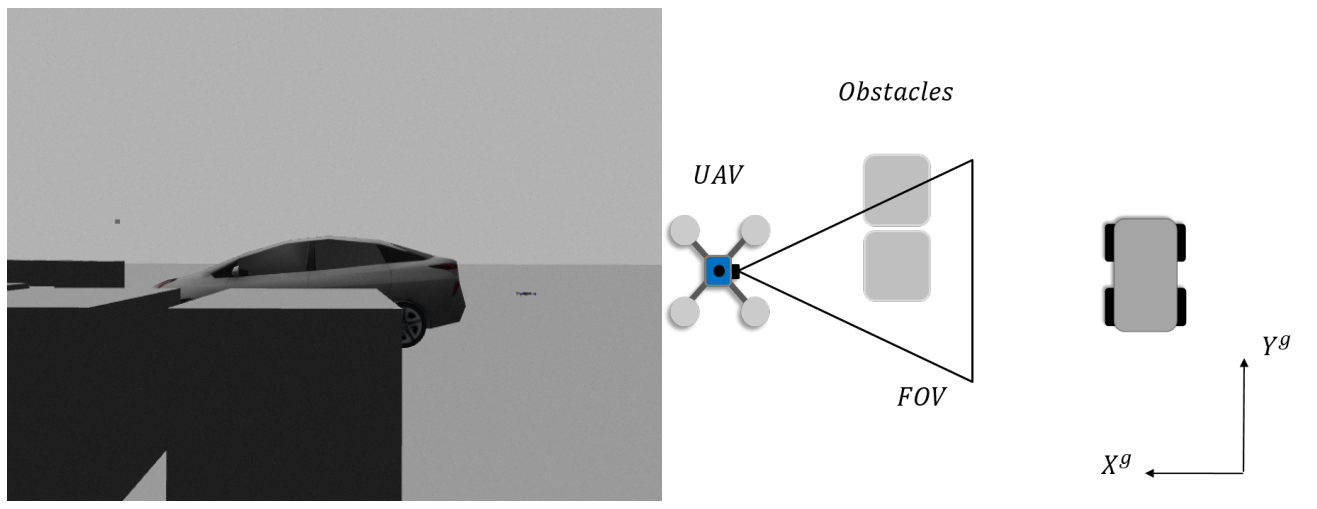

Fig. 7. Diagram of occlusion during the tracking mission.

During the tracking mission, the moving target in the image plane of the $i^{\text{th}}$ UAV might be occluded by static obstacles, leading to detection failure as shown in Fig. 7.

In Fig. 8, the position of each obstacle is defined as $p_o^j \in \mathbb{R}^3,\ j \in M \triangleq \{1,\ 2,\ \ldots,\ m\}$, and $n^i$ is defined as the vector from the $i^{\text{th}}$ UAV to the target in the global frame

$$n^i = (R^i)_c^g r_{q/c}^i, \tag{36}$$

where $(R^i)_c^g$ is the rotation matrix of the $i^{\text{th}}$ UAV from the camera frame to the global frame, and $r_{q/c}^i$ is defined in (1). To ensure that the $i^{\text{th}}$ UAV can always detect the target, occlusion angle $\theta^i \in \mathbb{R}$ obtained by vectors $p_o^j - p^i$ and $n^i$ needs to be kept larger than a user-defined FOV (field of view) $\theta^*$. In this way, it creates an occlusion-free area in the image, and the CBF for occlusion avoidance $h_{ij}^{o}$ can be designed as

$$h_{ij}^{o}(p^i, p_o^j) = \theta^i - \theta^* \geq 0, \tag{37}$$

where

$$\theta^i = \cos^{-1}\left( \frac{(p_o^j - p^i) \cdot n^i}{\left\| p_o^j - p^i \right\| \| n^i \|} \right). \tag{38}$$

The FOV $\theta^*$ considers the motion of the target and creates a moving occlusion-free area centered at the target in the image. Obstacles are allowed to partially show up in the image as long as they are not inside the occlusion-free area, as shown in Fig. 8, which can decrease the aggressiveness of the flight control input for the UAVs.

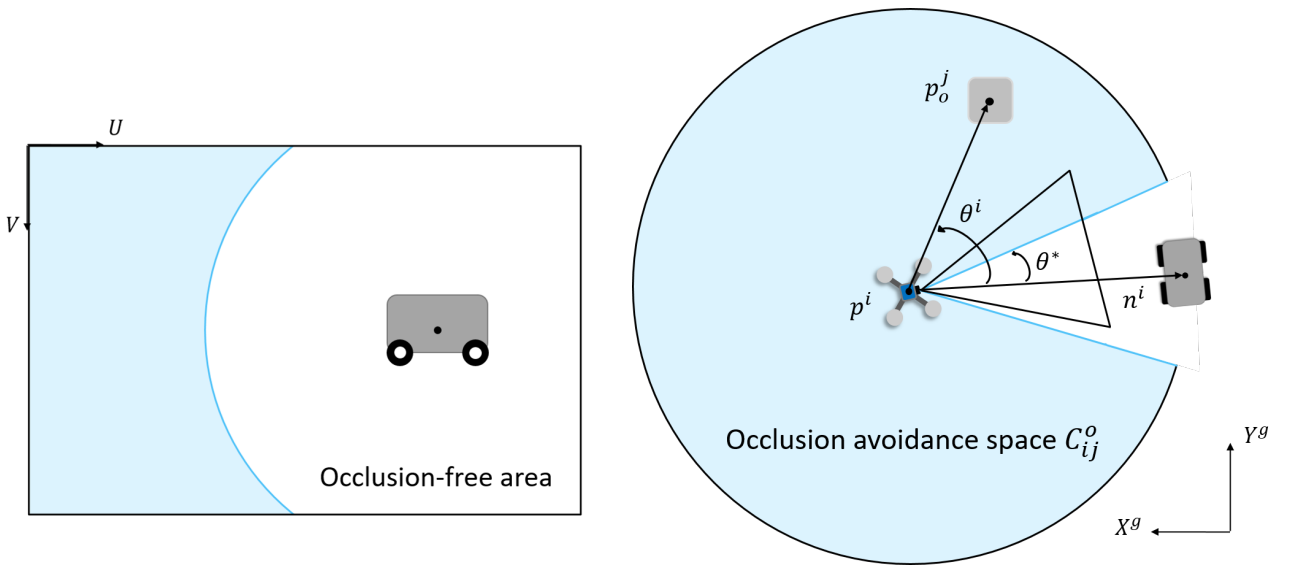

Fig. 8. Occlusion-free area and avoidance space. $n^i$ is the relative position between the target and the $i^{\text{th}}$ UAV. The blue zone is occlusion avoidance space $C^o_{ij}$ where the $j^{\text{th}}$ obstacle will not block the target in the image. The white zone is the occlusion-free area where the $j^{\text{th}}$ obstacle is not allowed to exist.

Combining (37) and (38), occlusion avoidance space $C^o_{ij},\ \forall i \in N,\ j \in N_{oi}$ can be defined as

$$C^o_{ij} =\Bigg\{ p^i \in \mathbb{R}^3 \mid h^o_{ij}(p^i, p^j_o) \triangleq \cos^{-1}\left(\frac{(p^j_o - p^i)\cdot n^i}{\left\|p^j_o - p^i\right\| \|n^i\|}\right) - \theta^* \geq 0 \Bigg\}, \tag{39}$$

where $N_{oi} = \left\{ j \in M \mid \| p^j_o - p^i \| \leq \| r^i_{q/c} \| \right\}$ is the set of obstacles around the $i^{\text{th}}$ UAV since it is possible that they will occlude the target. Function $h^o_{ij}$ is considered to be a CBF that satisfies

$$\dot{h}^o_{ij}(p^i, p^j_o) \geq -\gamma_o h^o_{ij}, \tag{40}$$

where $\gamma_o$ is also a user-defined constant for fine tuning invariant set $C^o_{ij}$.

With (39) and given that $\frac{\mathrm{d}}{\mathrm{d}x}\cos^{-1}(\boldsymbol{x}) = \frac{-1}{\sqrt{1-\boldsymbol{x}^2}}$, with $\boldsymbol{x} = \frac{(p^j_o - p^i)\cdot n^i}{\left\|p^j_o - p^i\right\| \|n^i\|}$, (40) expressed as

$$\begin{aligned}
\dot{h}^o_{ij} &= \frac{\partial cos^{-1}(\boldsymbol{x})}{\partial \boldsymbol{x}} \frac{\partial(\boldsymbol{x})}{\partial t} \\
&= \frac{-1}{\sqrt{1-(\boldsymbol{x})^2}} \left( \frac{n^i}{\left\|p^j_o - p^i\right\| \|n^i\|} \cdot (\dot{p}^j_o - \dot{p}^i) \right) \\
&+ \frac{-1}{\sqrt{1-(\boldsymbol{x})^2}} \left( \frac{-(p^j_o - p^i) cos\theta^i}{\left\|p^j_o - p^i\right\|^2} \cdot (\dot{p}^j_o - \dot{p}^i) \right) \\
&+ \frac{-1}{\sqrt{1-(\boldsymbol{x})^2}} \left( \frac{(p^j_o - p^i)}{\left\|p^j_o - p^i\right\| \|n^i\|} \cdot \dot{n}^i \right) \\
&+ \frac{-1}{\sqrt{1-(\boldsymbol{x})^2}} \left( \frac{-n^i cos\theta^i}{\|n^i\|^2} \cdot \dot{n}^i \right) \geq -\gamma_o h^o_{ij},
\end{aligned} \tag{41}$$

where $\dot{n}^i$ can be expressed as

$$\begin{aligned}
\dot{n}^i &= (\dot{R}^i)^g_c r^i_{q/c} + (R^i)^g_c \dot{r}^i_{q/c} \\
&= (\hat{\omega}^i_c)^g (R^i)^g_c r^i_{q/c} + (R^i)^g_c (V^i_q - V^i_c - \omega^i_c \times r^i_{q/c}),
\end{aligned} \tag{42}$$

provided that the obstacles are static (i.e., $\dot{p}^j_o = 0$), where $\hat{(\cdot)}$ denotes the hat map, and $\dot{r}^i_{q/c}$ is defined in (2). Based on Assumption 2 (i.e., $V^i_q \approx V^i_c$), (42) can be further expressed as

$$\begin{aligned}
\dot{n}^i &= (\hat{\omega}^i_c)^g (R^i)^g_c r^i_{q/c} - (R^i)^g_c (\omega^i_c \times r^i_{q/c}) \\
&= (\hat{\omega}^i_c)^g n^i - (\omega^i_c)^g \times n^i \\
&= 0.
\end{aligned} \tag{43}$$

Based on (43), (41) can be rewritten as

$$\begin{aligned}
\dot{h}^o_{ij} &= \frac{-1}{\sqrt{1-(\boldsymbol{x})^2}} \frac{n^i}{\left\|p^j_o - p^i\right\| \|n^i\|} (V^i_c)^g \\
&- \frac{-1}{\sqrt{1-(\boldsymbol{x})^2}} \frac{(p^j_o - p^i) cos\theta^i}{\left\|p^j_o - p^i\right\|^2} (V^i_c)^g \leq \gamma_o h^o_{ij},
\end{aligned} \tag{44}$$

which can be transformed into a linear inequality

$$A^o_{ij}(u^i)^g \leq b^o_{ij}\ \forall i \in N,\ j \in N_{oi},$$

where

$$\begin{aligned}
A^o_{ij} =& \Bigg[ \underbrace{\frac{-1}{\sqrt{1-(\boldsymbol{x})^2}} \left( \frac{n^i}{\left\|p^j_o - p^i\right\| \|n^i\|} - \frac{(p^j_o - p^i) cos\theta^i}{\left\|p^j_o - p^i\right\|^2} \right)^{\mathrm{T}}}_{i^{\text{th}}\ \text{UAV}} \\
&, \ 0^{1\times 3} \Bigg] \\
b^o_{ij} =& \gamma_o h^o_{ij}.
\end{aligned} \tag{45}$$

In other words, the occlusion of the target is avoided if control input $(u^i)^g$ stays in admissible control space $K^o_{ij}$ defined as

$$K^o_{ij} = \left\{ (u^i)^g \in \mathbb{R}^6 \mid A^o_{ij}(u^i)^g \leq b^o_{ij}\ \forall i \in N,\ j \in N_{oi} \right\}. \tag{46}$$

## V. Controller Design

To control both the relative positions and angles between the target and the team of UAVs to be within desired values, the objective of this work is to control the UAVs to achieve

$$s^i_m \to s^{i*}.$$

To guarantee the cooperative tracking by the UAVs is feasible, constraints of their interaction and the generated control inputs need to be enforced. Therefore, the following optimal control problem is defined.

Given optimal tracking control input $u^i$ obtained from the NMPC designed in (18), its augmented controller $\hat{u}^i = (R^i)^g_c \left[ v^i_{cx}, v^i_{cy}, v^i_{cz}, 0, \omega^i_{cy}, 0 \right]^{\mathrm{T}} \in \mathbb{R}^6$ expressed in the global frame is defined. Then controller $(u^i)^g$, defined as a minimally invasive controller of $\hat{u}^i$, is solved as the following optimization problem using standard QP solvers to achieve the tracking performance while ensuring that the closed-loop system remains in the forward invariant set of the CBFs defined in Section IV:

$$\min_{(u^i)^g \in \mathbb{R}^6} \parallel (u^i)^g - \hat{u}^i \parallel^2, \tag{47}$$

$$\begin{aligned} s.t.\ & A^s_{ij}(u^i)^g + s_s = b^s_{ij}, \forall i \in N,\ j \in N_{si}, \\ & A^c_{ij}(u^i)^g + s_c = b^c_{ij},\ \forall i \in N,\ j \in N_{ci}, \\ & A^o_{ij}(u^i)^g + s_o = b^o_{ij},\ \forall i \in N,\ j \in N_{oi}, \\ & \parallel (V^i_c)^g \parallel \leq \alpha_v,\ \forall i \in N, \\ & \parallel (\omega^i_c)^g \parallel \leq \alpha_\omega,\ \forall i \in N, \\ & \frac{4\alpha_v R_s}{D^2_s - R^2_s} \leq \gamma_s,\ \forall i \in N, \\ & s_s, s_c, s_o \geq 0, \end{aligned}$$

where the constraints are defined in (25), (34), and (45), $s_s, s_c, s_o$ are slack variables, and $\alpha_\omega$ is the upper bound of the angular velocities of the camera. The controller architecture is depicted in Fig. 9, and the performance of the controller is demonstrated by the simulations described in Section VI.

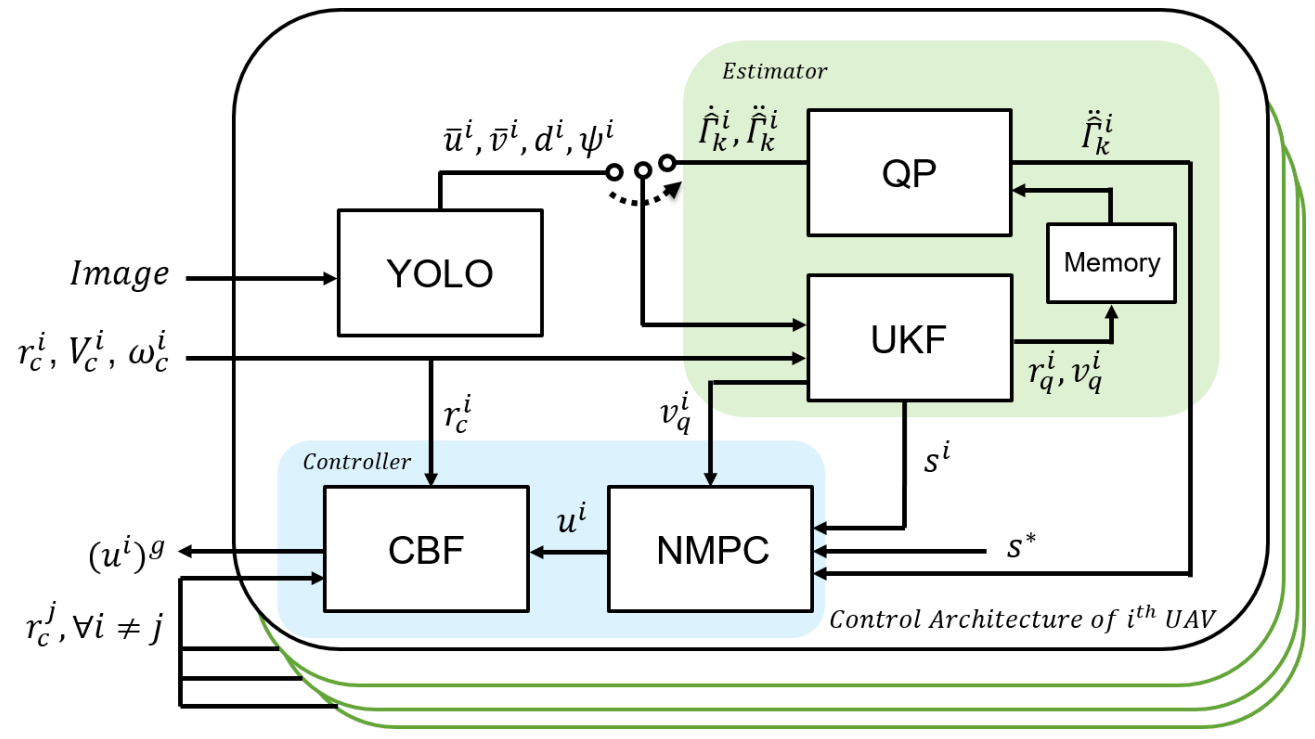

Fig. 9. Block diagram of the tracking controller that includes consideration of CBFs.

*Remark* 6. Based on the first three constraints of the optimization problem defined in (47), position of the UAV is communicated among the agents for feedback of the distributed controller. Furthermore, the constraints of the optimal problem defined in (48) are derived based on CBFs, and the barrier functions satisfying these conditions imply forward invariance of the set based on [21]. Therefore, as long as the constraints are initially satisfied, it will remain inside the forward invariance set, and the control objective can be achieved.

## VI. Simulations

To demonstrate the efficacy of the developed tracking controller, three UAVs with different initial conditions were utilized to track a target moving in three cluttered environment.

### A. Environment Setup

Simulations were conducted using the Gazebo simulator in the ROS framework (version 18.04, Melodic). The simulations included UAV1, UAV2, and UAV3 with initial locations of $\begin{bmatrix} 0, & 6, & 0.85 \end{bmatrix}^{\mathrm{T}}$, $\begin{bmatrix} -6, & 0, & 0.85 \end{bmatrix}^{\mathrm{T}}$, and $\begin{bmatrix} 0, & -6, & 0.85 \end{bmatrix}^{\mathrm{T}}$, and initial orientations of $\begin{bmatrix} 0°, & 0°, & -90° \end{bmatrix}^{\mathrm{T}}$, $\begin{bmatrix} 0°, & 0°, & 0° \end{bmatrix}^{\mathrm{T}}$,and $\begin{bmatrix} 0°, & 0°, & 90° \end{bmatrix}^{\mathrm{T}}$ in the global frame ($X^g$-$Y^g$-$Z^g$) in scenario A as shown in Fig. 10. The reference feature vectors are prescribed as $s^{1,*} = [0, 0.188, 1/7, \pi/2]$, $s^{2,*} = [0, 0.188, 1/7, 0]$, and $s^{3,*} = [0, 0.188, 1/7, -\pi/2]$, and FOV $\theta^* = 30^\circ$ The target was located at $\begin{bmatrix} 0, & 0, & 0 \end{bmatrix}^{\mathrm{T}}$ with an initial orientation of $\begin{bmatrix} 0°, & 0°, & 0° \end{bmatrix}^{\mathrm{T}}$ in the global frame. When the target begins to move, the UAVs track the target and estimate its position and velocity. The ground-truth dimensions of the target were 4.6 m × 1.8 m × 1.5 m (length × width × height). Ten boxes of size 1 m × 1 m × 1 m lined up alongside the path were used as obstacles, as shown in Fig. 10. The positions of the static obstacles were assumed to be known, and the target was free to move in the $X^g$-$Y^g$ plane, with its velocity specified by user commands.

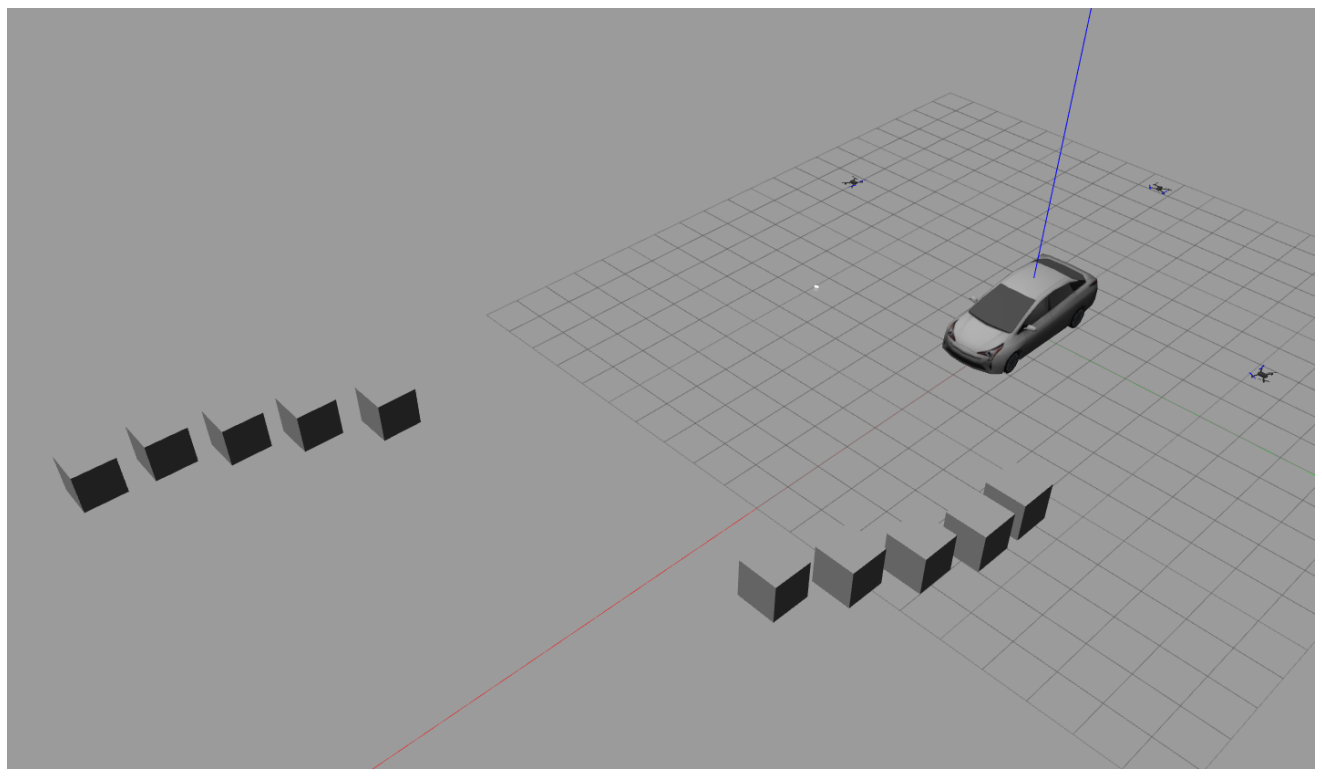

Fig. 10. Scenario A: Initial setup of a car and three UAVs.

The obtained images had a resolution of 640 × 480, and the intrinsic parameters matrix of the camera was

$$K = \begin{bmatrix} 381.36 & 0 & 320 \\ 0 & 381.36 & 240 \\ 0 & 0 & 1 \end{bmatrix},$$

as obtained by calibration. The simulation did not require prior knowledge about the moving target, and the sampling rate was fixed at 80 Hz for the detection, state estimation, noise estimation, and control processes. In order to maintain the visibility and address the limitations of the actuator while tracking the moving target, the constraints of the states and constraints of the control inputs were considered in the numerical simulations, as listed in Table I and II. The parameters for tuning the sampling time, predicted horizons, and the weighting matrices in the cost function are presented in Table III. The parameters for deciding the safety ranges and the restrictiveness of the control space in CBFs are listed in Table IV.

TABLE I
STATE CONSTRAINTS.

| State | Minimum | Maximum |
|---|---|---|
| $x_1^i$ | $-0.84$ | $0.84$ |
| $x_2^i$ | $-0.63$ | $0.63$ |
| $x_3^i$ | $0.07$ | $1$ |
| $\psi^i$ | $0^\circ$ | $180^\circ$ |

TABLE II
CONTROL CONSTRAINTS.

| Inputs | Minimum | Maximum |
|---|---|---|
| $v_{cx}^i, v_{cy}^i, v_{cz}^i$ (m/s) | $-10$ | $10$ |
| $w_{cy}^i$ (rad/s) | $-0.6$ | $0.6$ |

TABLE III
CONTROL PARAMETER SETTINGS.

| Parameter | Value |
|---|---|
| Horizon length ($N_p$) | 50 |
| $Q_s$ | $\begin{bmatrix} 1 & 1 & 1 & 1 \\ 1 & 1 & 1 & 1 \\ 1 & 1 & 100 & 1 \\ 1 & 1 & 1 & 1 \end{bmatrix}$ |
| $R_u$ | $\begin{bmatrix} 0.02 & 1 & 1 & 1 \\ 1 & 0.03 & 1 & 1 \\ 1 & 1 & 0.01 & 1 \\ 1 & 1 & 1 & 0.3 \end{bmatrix}$ |

TABLE IV
CBF PARAMETER SETTINGS.

| Parameter | Value | Parameter | Value |
|---|---|---|---|
| $R_s$ (m) | 2 | $\gamma_s$ | 3 |
| $R_c$ (m) | 20 | $\gamma_c$ | 1 |
| $\theta^\circ$ (deg) | 30 | $\gamma_o$ | 0.1 |
| $D_s$ (m) | 20 | $\alpha_v$ (m/s), $\alpha_\omega$ (rad/s) | 10, 0.6 |

The simulation scenario A is depicted in Fig. 11. With the aim of collision avoidance and occlusion avoidance for obstacles, static obstacles numbered from 1 to 5 and from 6 to 10 were placed along the route where UAV1 and UAV3 were to track the moving target, receptively.

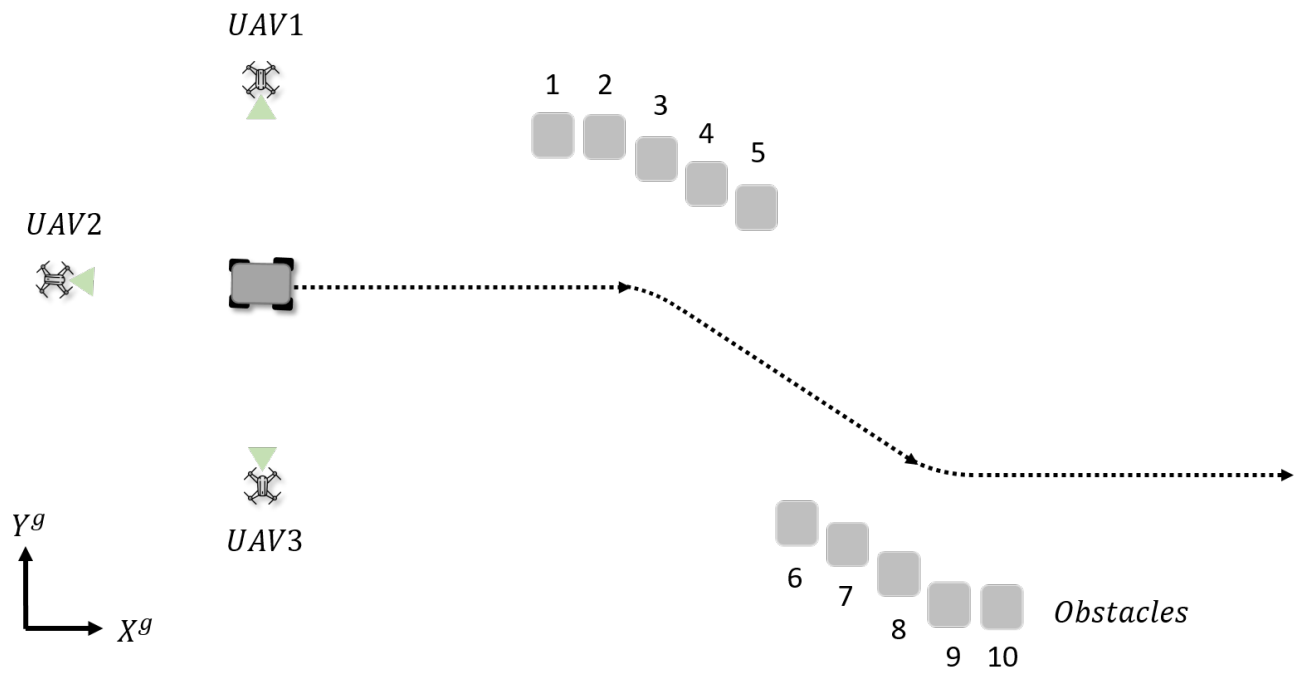

Fig. 11. Simulation scenario A. UAV1, UAV2, and UAV3 tracked a target moving over a z-shaped path. The trajectory of the target was unknown to the UAVs. The target moved at 0.5 m/s in the heading direction from 0 s to 10 s, and first turned at $-0.1$ m/s$^2$ from 10 s to 15 s. The target kept moving at 0.5 m/s from 15 s to 35 s, and performed a second turn at 0.1 m/s$^2$ from 35 s to 40 s, and then kept moving at 0.5 m/s.

## B. Simulation Results

*a) State Feature Vectors of the UAVs:* Fig. 12 and 13 show that the UAVs kept the moving target within the FOV of the cameras at the desired depth in this simulation.

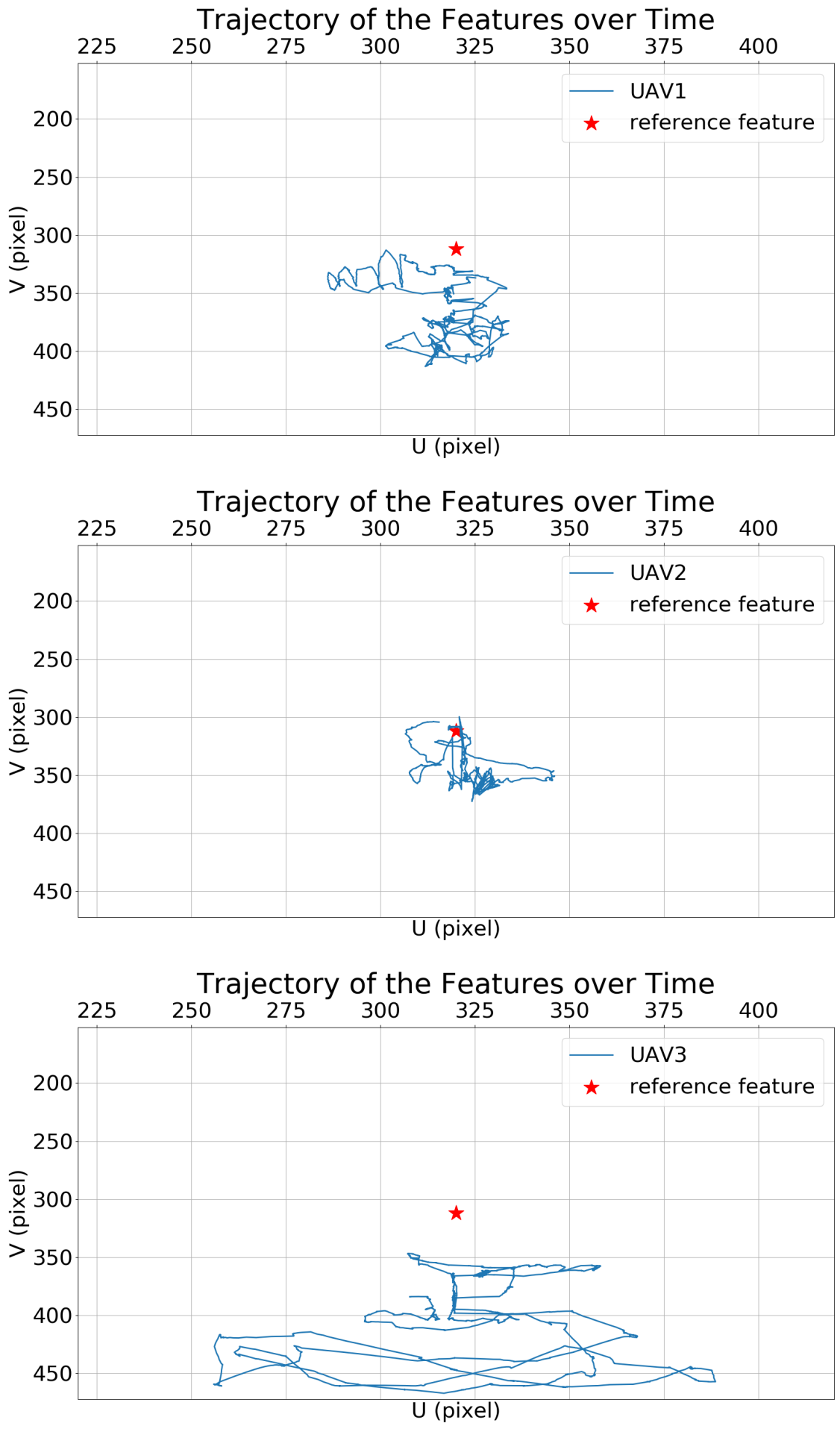

Fig. 12. Trajectory of the moving target in the image frame observed by UAVs.

The performance in Fig. 12 is quantified as the root mean square (RMS) tracking errors in Table V, where the values denote the position errors between the center of the bounding box and the reference location (i.e., $[x_1^*, x_2^*]$ defined in (17) ) in the image frame. However, the baseline RMS errors are hard to obtain since the tracking errors can vary significantly in different scenarios and initial conditions.

TABLE V
RMS ERRORS IN IMAGE FEATURES.

| Agent | $U$-axis | $V$-axis | Unit |
|---|---|---|---|
| UAV1 | 10.26 | 63.23 | pixel |
| UAV2 | 8.94 | 32.66 | pixel |
| UAV3 | 23.22 | 86.28 | pixel |

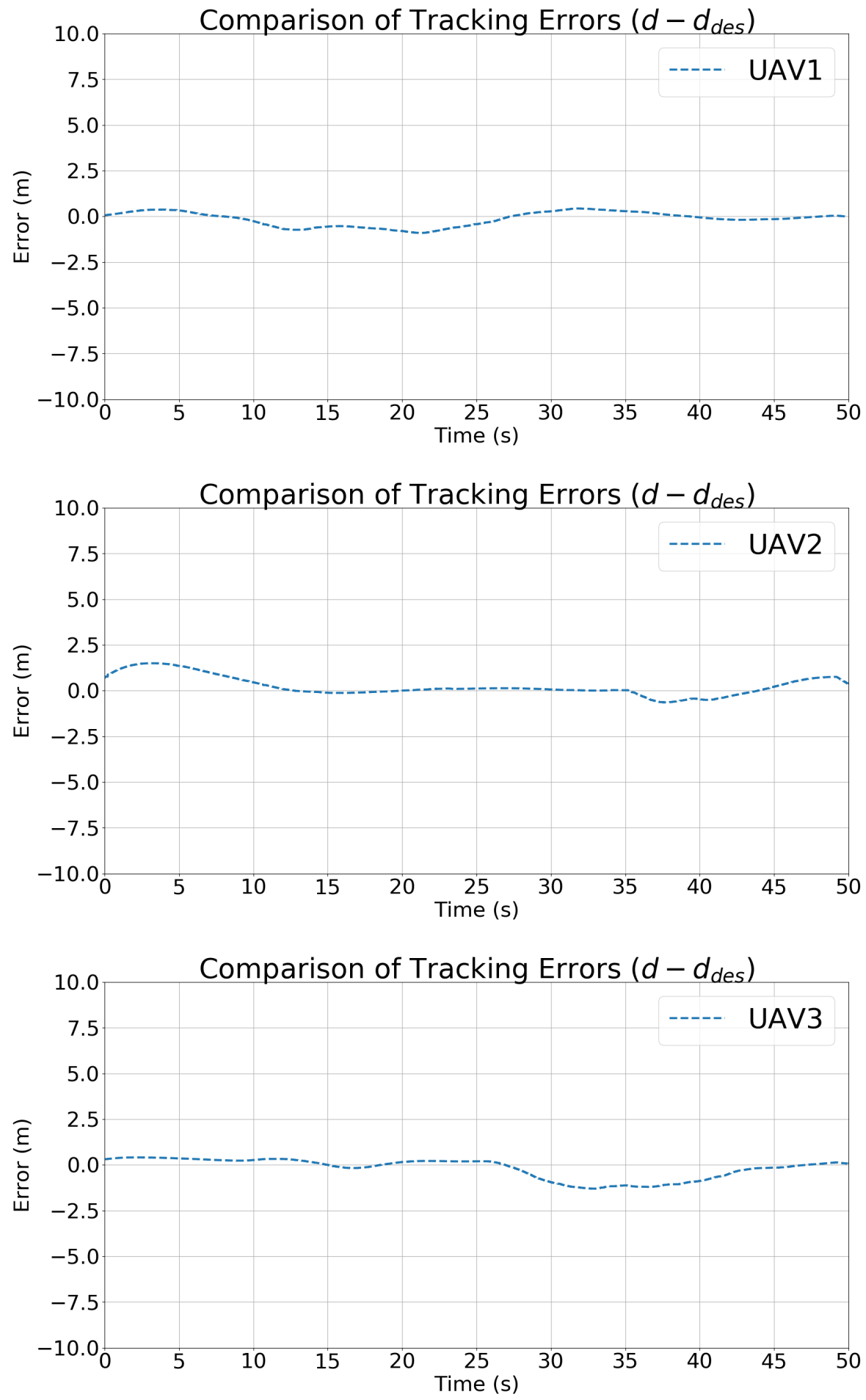

Fig. 13. Tracking errors of depth for UAVs.

*b) Distances Between UAVs :* Fig. 14 shows the distances between the UAVs. This figure indicates that the constraints of connectivity maintenance and collision avoidance were satisfied among the UAVs, since they remained within connectivity $R_c = 20$ m and collision range $R_s = 2$ m.

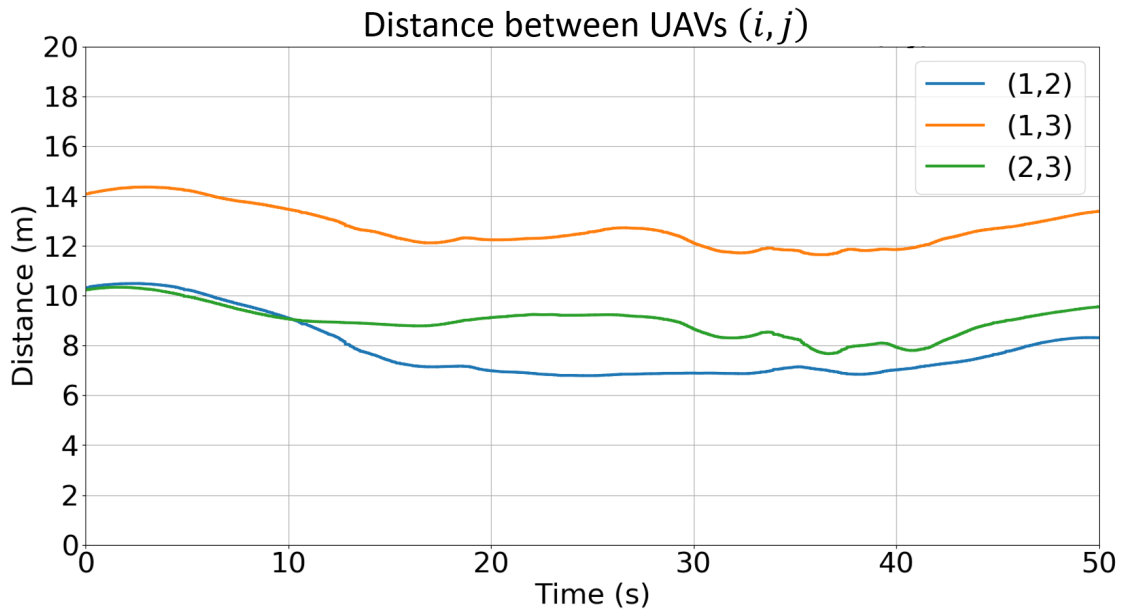

Fig. 14. Distances between the UAVs

*c) Distances Between UAVs and Obstacles:* Fig. 15 and 16 show the distances from the obstacles to UAV1 and UAV3, respectively. These figures indicate that the constraints between the obstacles and UAVs were also satisfied since they followed the rule of keeping a safe distance $R_s = 2$ m from each other. UAV1 avoided colliding with obstacles 1 to 5 around the time period from 10 s to 30 s, while UAV3 avoided obstacles 6 to 10 from around 30 s to 50 s. Since UAV2 did not encounter any obstacles, its distances are not shown.

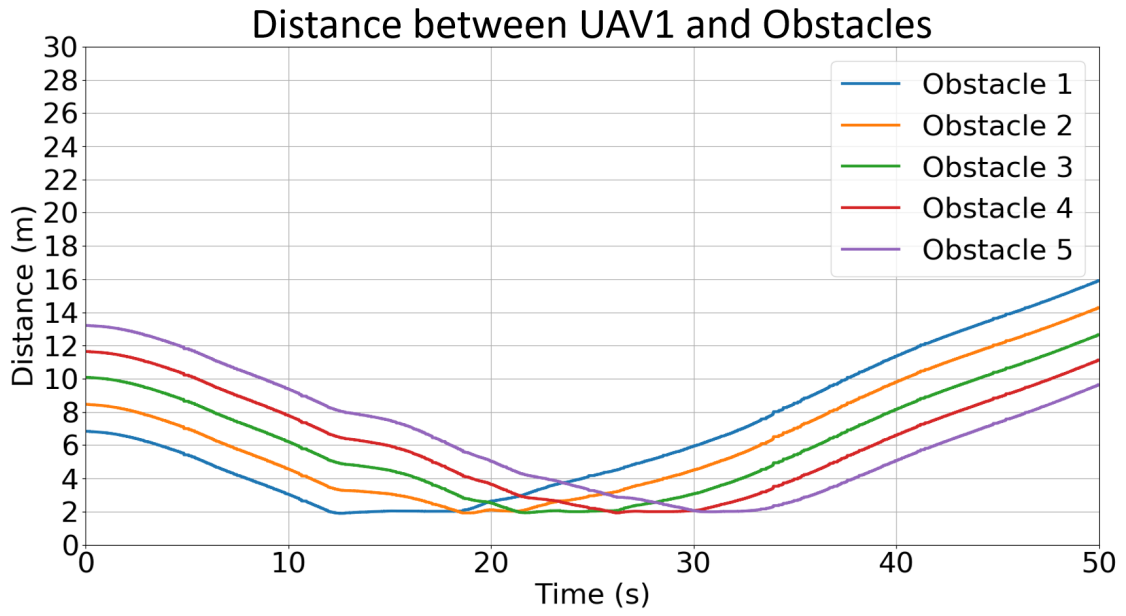

Fig. 15. Distance between UAV1 and the obstacles.

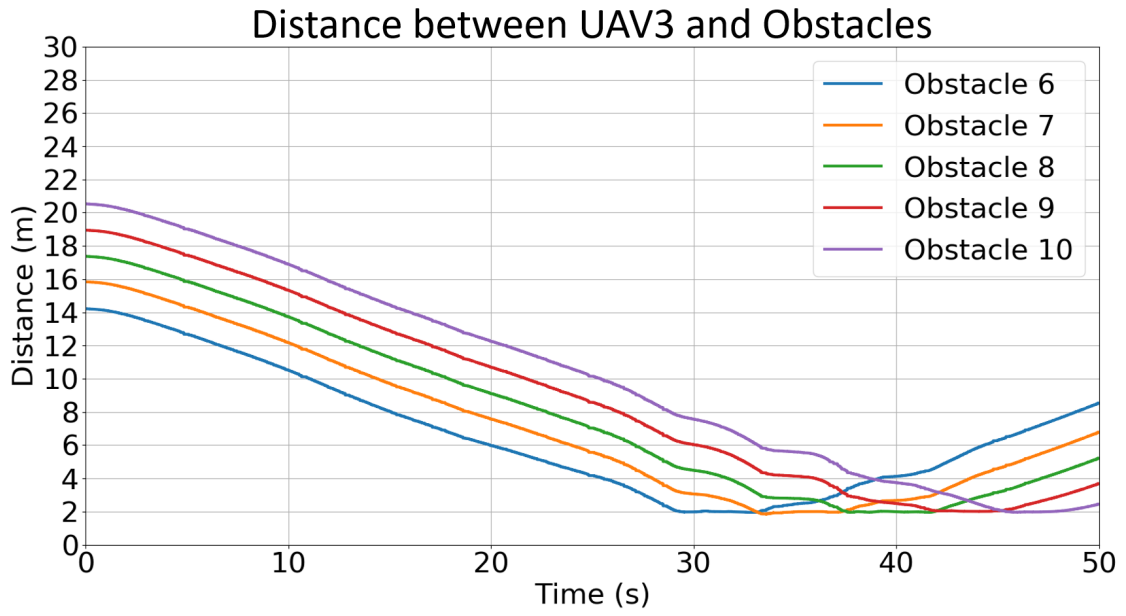

Fig. 16. Distance between UAV3 and the obstacles.

*d) Trajectories of UAVs :* The left figure in Fig. 17 shows the 2D trajectories of the UAVs in the $X^g$-$Y^g$ plane. This figure indicates that when UAV1 first approached obstacle 1, it moved forward to the obstacle to prevent the obstacle from showing up in the camera image; UAV3 similarly did that when it approached obstacle 6. The right figure in Fig. 17 shows the 2D trajectories of the UAVs in the $X^g$-$Z^g$ plane. This figure indicates that when UAV1 approached obstacle 1, it moved upward to avoid colliding with the obstacles while also ensuring occlusion avoidance. UAV1 moved downward when it left the obstacles in order to keep the target within the FOV of the camera and thereby allow it to be tracked. The figure shows that UAV3 also satisfied the aforementioned behavior.

*e) Occlusion Angles Observed by UAVs:* Fig. 18 and 19 show the occlusion angles observed by UAV1 and UAV3 and the slack variables. These figures indicate that occlusion was avoided between the obstacles and the UAVs. UAV1 avoided occlusion when it approached obstacles 1 to 5 from 10 s to 30 s, and UAV3 avoided occlusion when it approached obstacles 6 to 10 from 30 s to 50 s. Furthermore, all slack variables satisfy the constraints (i.e., greater than 0).

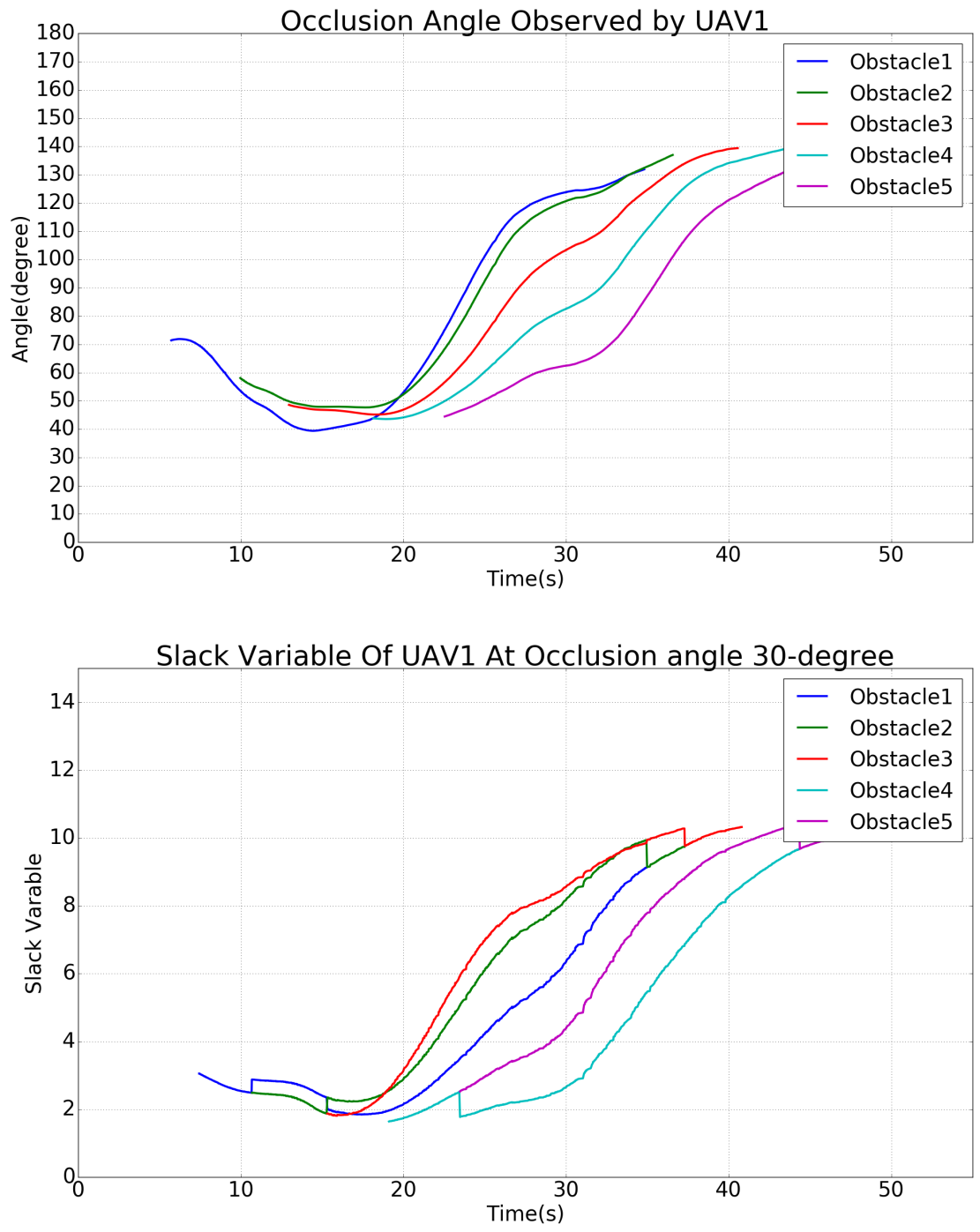

Fig. 18. Occlusion angles observed by UAV1 (upper) and the slack variables (lower).

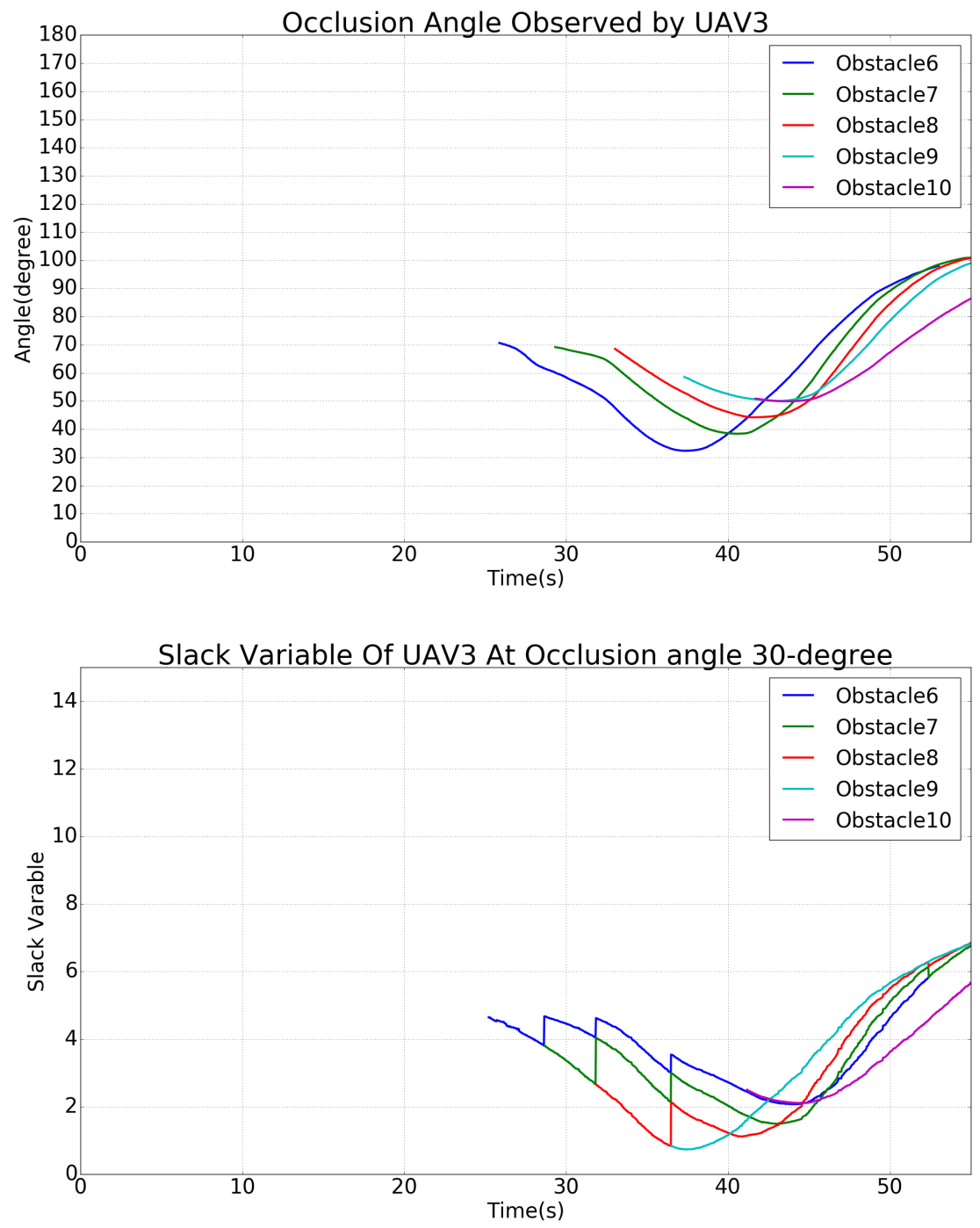

Fig. 19. Occlusion angles observed by UAV3 for $\gamma_o = 0.1$ (upper) and the slack variables (lower).

*f) Performance in Occlusion Avoidance :* The choice of $\gamma_o$ can change the performance of the developed controller in occlusion avoidance. A higher $\gamma_o$ results in a more restrictive occlusion avoidance space, as defined in (40). The performance in occlusion avoidance was better with $\gamma_o = 0.5$ (Fig. 20) than with $\gamma_o = 0.1$ (Fig. 19). However, a higher $\gamma_o$ will also lead to more aggressive and less smooth motion in order to meet the more restrictive constraints.

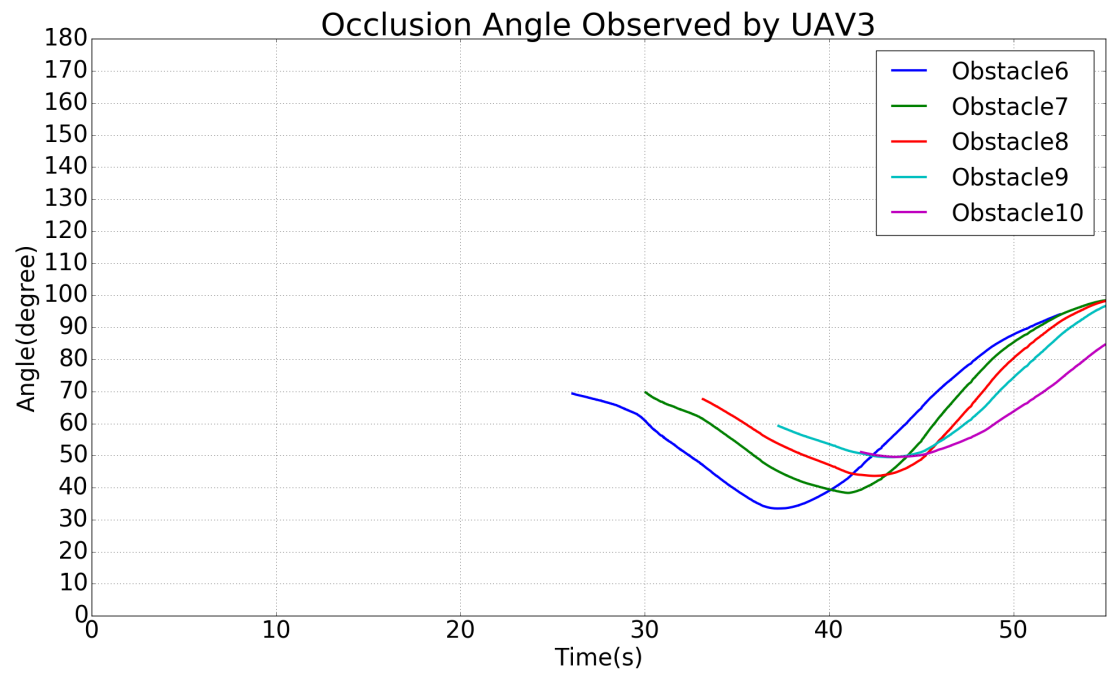

Fig. 20. Occlusion angles observed by UAV3 for $\gamma_o = 0.5$.

## *C. Simulation in Scenarios B and C*

To further verify the efficacy of the developed controller, simulations in scenarios B and C with different initial conditions as shown in Figs. 21 and 22 are conducted to evaluate the performance of the system. The simulation results are summarized in Table VI. Two initial conditions are set in each scenario, and all the constraints are satisfied. The control objectives are achieved with different tracking errors, and larger tracking errors are observed in V-axis in scenario C due to the higher obstacles.

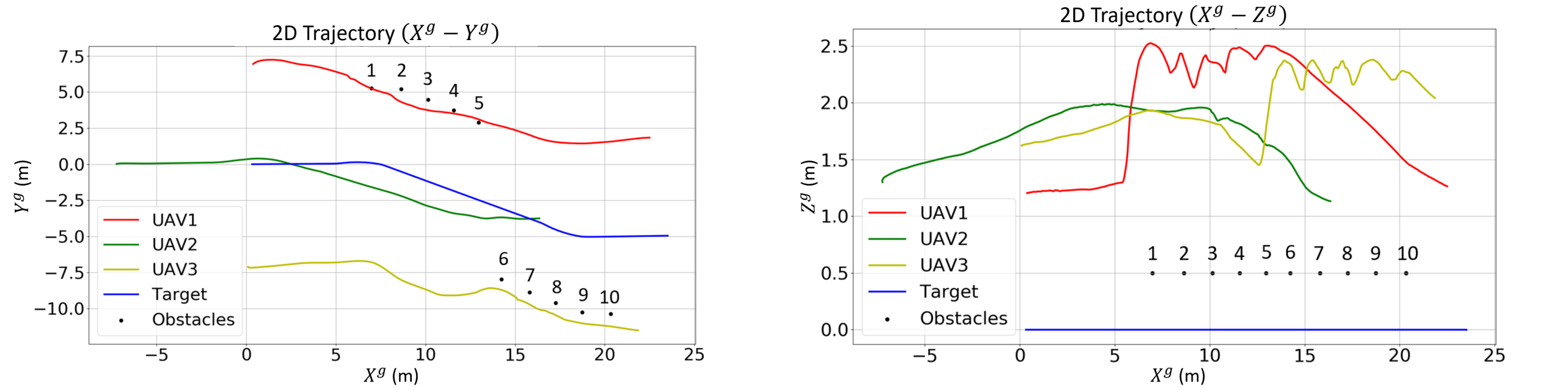

Fig. 17. 2D trajectories of the UAVs in the $X^g$-$Y^g$ plane (Left) and 2D trajectories of the UAVs in the $X^g - Z^g$ plane (Right).

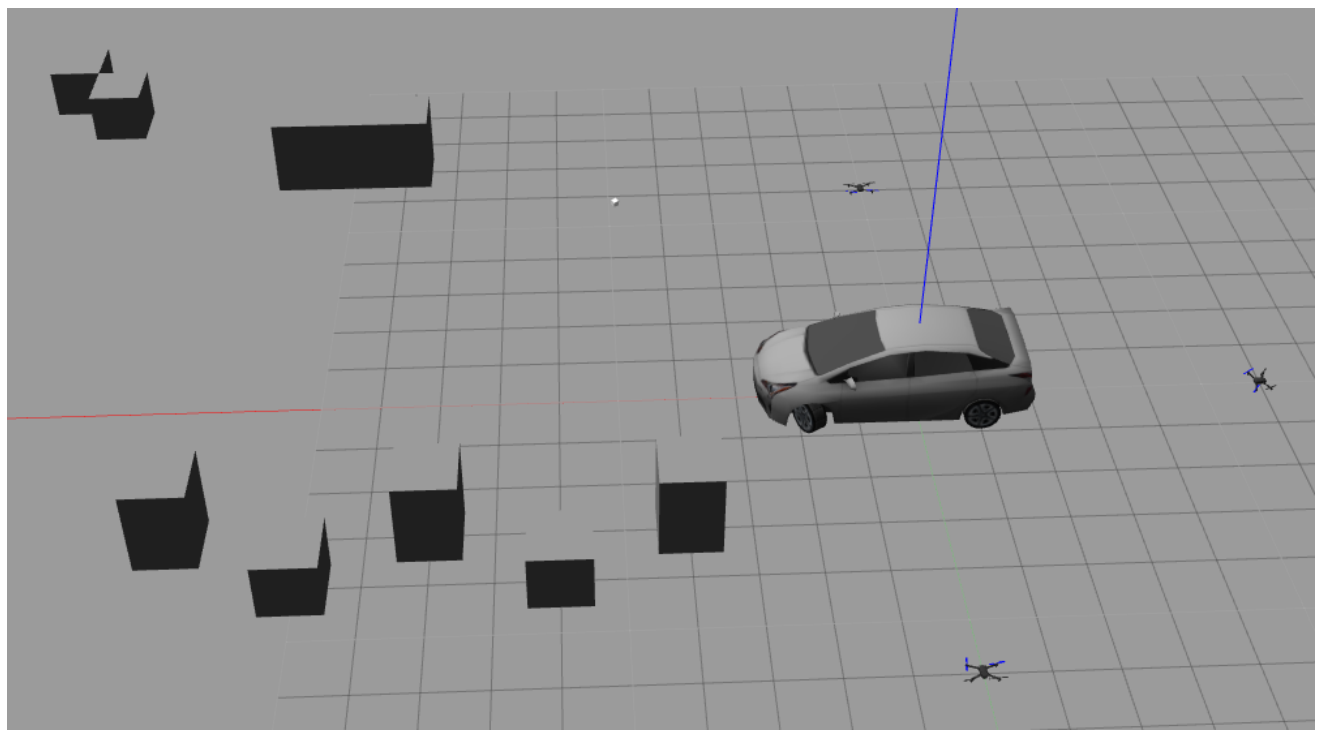

Fig. 21. Scenario B

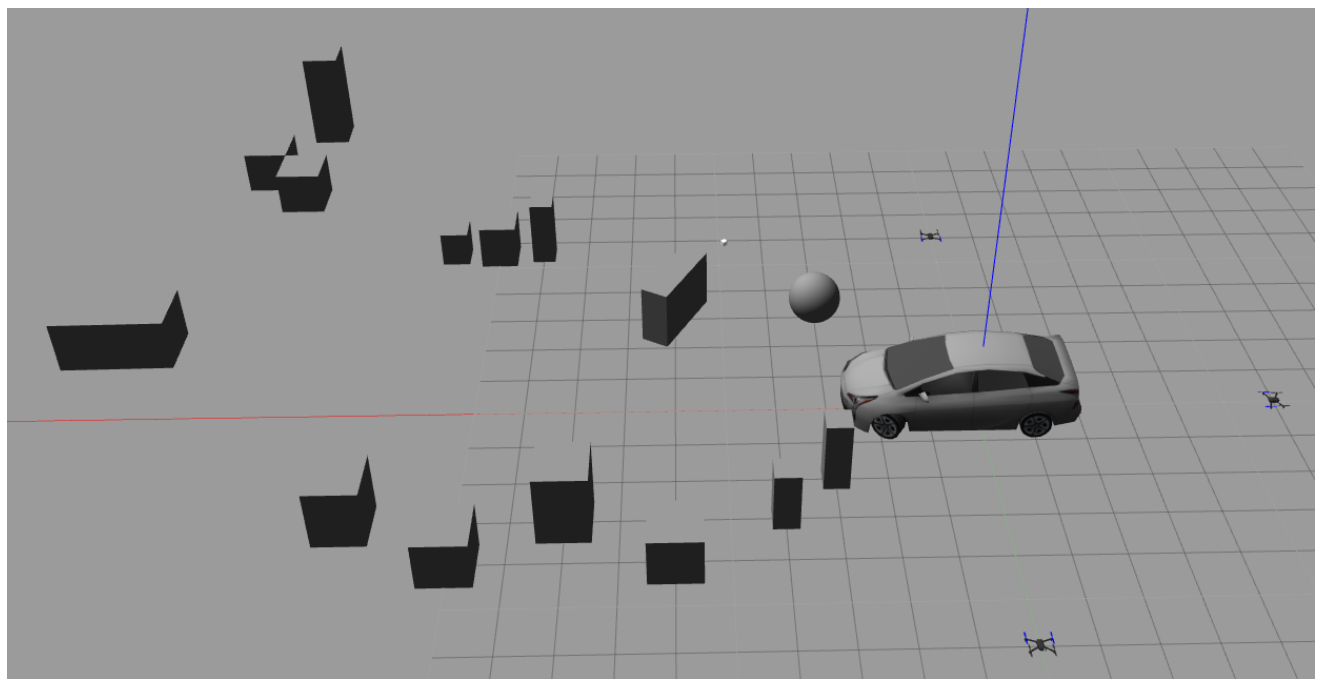

Fig. 22. Scenario C

## VII. Conclusions

The simulation results presented here indicate that the developed CBFs can guarantee the safety of a multi-UAV tracking system in a cluttered environment. The modified controller solved based on the NMPC tracking controller using QP can generate minimally invasive control actions and meet the following objectives:

- All UAVs are proven to be safe since the constraints ensure that collisions are prevented, connectivity is maintained, and occlusion of the target in the camera image is avoided.
- The constraints are designed based on information about the neighboring UAVs, which decreases the communication burden and improves the performance of the controller.
- Relative angle $\psi$ can be prescribed by the user for the desired angle of tracking and observations.

One of the contributions of this work is the relative angle $\psi$ can be measured by the YOLO DNN, which enables the UAVs to track the target from the desired relative angle to the target. This capability is important in many applications. For example, to autonomously check the license plate of suspect vehicles, the UAV needs to fly to the front or back part of the vehicle based on relative angle feedback. Therefore, the relative angle is highly related to the aim of the task. This capability can also be extended to many applications such as formation, surveillance, etc. In addition, increasing coefficient $\gamma_o$ will make the occlusion avoidance space more restrictive but will also increase the aggressiveness of the control input. This implies that the approximation error to the NMPC controller defined in (47) is compromised in order to meet the occlusion avoidance constraint. However, gimbals were not included in the current system, and new motion patterns of the UAVs can be expected once a gimbal with two to three degrees of freedom is installed in each UAV. The inclusion of a gimbal in each UAV will facilitate observations of the target with different attitudes at the same location, which implies that their accelerations would also be different. Future works could assume that the positions of the obstacles are unknown and so need to be detected using other algorithms, such as SLAM (simultaneous localization and mapping). Additionally, a CBF for occlusion avoidance could be developed that considers not only static obstacles but also moving obstacles such as birds, humans, and bicycles. Moreover, investigating the interplay the air turbulence among the UAVs can also improve the instability of the UAVs.

## References

[1] J. Valasek, K. Kirkpatrick, J. May, and J. Harris, "Intelligent motion video guidance for unmanned air system ground target surveillance," *J. Aerosp. Inf. Syst.*, vol. 13, no. 1, pp. 10–26, 2016.

TABLE VI
SIMULATIONS RESULTS OF SCENARIOS B AND C

| Scenarios | UAV | Initial Condition (x, y, heading angle) | Occlusion Avoidance | Connectivity Maintenance | Collision Avoidance | RMS Tracking Error U-axis (pixel) | RMS Tracking Error V-axis (pixel) |
|---|---|---|---|---|---|---|---|
| A | 1 | (0, 6, -1.57) | Y | Y | Y | 10.26 | 63.23 |
| | 2 | (-6, 0, 0) | Y | Y | Y | 8.94 | 32.66 |
| | 3 | (0, -6, 1.57) | Y | Y | Y | 23.22 | 86.28 |
| | 1 | (0, 6, -1.57) | Y | Y | Y | 13.44 | 54.04 |
| | 2 | (-6, 3, 0) | Y | Y | Y | 11.48 | 40.38 |
| | 3 | (-3, -6, 1.57) | Y | Y | Y | 12.46 | 44.05 |
| B | 1 | (0, 6, -1.57) | Y | Y | Y | 44.21 | 82.76 |
| | 2 | (-6, 0, 0) | Y | Y | Y | 32.97 | 71.51 |
| | 3 | (0, -6, 1.57) | Y | Y | Y | 28.88 | 72.98 |
| | 1 | (-4, 5, -0.784) | Y | Y | Y | 36.77 | 86.23 |
| | 2 | (-8, 0, 0) | Y | Y | Y | 28.49 | 67.86 |
| | 3 | (-3, -6, 0.784) | Y | Y | Y | 24.67 | 68.32 |
| C | 1 | (0, 6, -1.57) | Y | Y | Y | 36.68 | 140.38 |
| | 2 | (-6, 0, 0) | Y | Y | Y | 51.95 | 146.17 |
| | 3 | (0, -6, 1.57) | Y | Y | Y | 43.72 | 130.31 |
| | 1 | (-2, 6, -1.2) | Y | Y | Y | 34.69 | 110.17 |
| | 2 | (-6, 0, 0) | Y | Y | Y | 18.37 | 92.31 |
| | 3 | (-2, -6, 1.2) | Y | Y | Y | 24.99 | 100.15 |

[2] S. Papatheodorou, A. Tzes, and Y. Stergiopoulos, "Collaborative visual area coverage," *Robot. Autonom. Syst.*, vol. 92, pp. 126 – 138, 2017.

[3] T. Nï¿œgeli, J. Alonso-Mora, A. Domahidi, D. Rus, and O. Hilliges, "Real-time motion planning for aerial videography with dynamic obstacle avoidance and viewpoint optimization," *IEEE Robot. Autom. Lett.*, vol. 2, no. 3, pp. 1696–1703, 2017.

[4] R. Bonatti, Y. Zhang, S. Choudhury, W. Wang, and S. A. Scherer, "Autonomous drone cinematographer: Using artistic principles to create smooth, safe, occlusion-free trajectories for aerial filming," *CoRR*, vol. abs/1808.09563, 2018.

[5] A. Macwan, G. Nejat, and B. Benhabib, "Target-motion prediction for robotic search and rescue in wilderness environments," *IEEE Trans. Syst. Man Cybern., Part B (Cybern.)*, vol. 41, no. 5, pp. 1287–1298, Oct 2011.

[6] J.-M. Li, C.-W. Chen, and T.-H. Cheng, "Motion prediction and robust tracking of a dynamic and temporarily-occluded target by an unmanned aerial vehicle," *IEEE Trans. Control Syst. Technol.*, pp. 1–13, 2020, dOI:10.1109/TCST.2020.3012619.

[7] R. Tallamraju, E. Price, R. Ludwig, K. Karlapalem, H. H. Bï¿œlthoff, M. J. Black, and A. Ahmad, "Active perception based formation control for multiple aerial vehicles," *IEEE Robotics and Automation Letters*, vol. 4, no. 4, pp. 4491–4498, 2019.

[8] Y. Zhao, X. Wang, C. Wang, Y. Cong, and L. Shen, "Systemic design of distributed multi-uav cooperative decision-making for multi-target tracking," *Auton. Agents Multi-Agent Syst.*, vol. 33, no. 1-2, pp. 132–158, 1 2019.

[9] M. Razzanelli, M. Innocenti, G. Pannocchia, and L. Pollini, *Vision-based Model Predictive Control for Unmanned Aerial Vehicles Automatic Trajectory Generation and Tracking*.

[10] I. F. Mondragï¿œn, P. Campoy, M. A. Olivares-Mendez, and C. Martinez, "3D object following based on visual information for unmanned aerial vehicles," in *Proc. IEEE IX Latin Am. Robot. Symp. IEEE Colomb. Conf. Autom. Control*, Oct 2011, pp. 1–7.

[11] H. Wang, B. Yang, J. Wang, X. Liang, W. Chen, and Y. Liu, "Adaptive visual servoing of contour features," *IEEE/ASME Trans. Mechatron.*, vol. 23, no. 2, pp. 811–822, Apr. 2018.

[12] D. Hulens and T. Goedemï¿œ, "Autonomous flying cameraman with embedded person detection and tracking while applying cinematographic rules," in *Conf. Comput. Robot Vis.*, Edmonton, AB, Canada, May 2017, pp. 56–63.

[13] Y. Liu, Q. Wang, H. Hu, and Y. He, "A novel real-time moving target tracking and path planning system for a quadrotor uav in unknown unstructured outdoor scenes," *IEEE Trans. Syst., Man, Cybern., Syst.*, vol. 49, no. 11, pp. 2362–2372, Nov 2019.

[14] J. H. Lee, J. D. Millard, P. C. Lusk, and R. W. Beard, "Autonomous target following with monocular camera on uas using recursive-ransac tracker," in *Int. Conf. Unmanned Aircr. Syst.*, Dallas, TX, USA, Jun. 2018, pp. 1070–1074.

[15] T. P. Nascimento, L. F. S. Costa, A. G. S. Conceiï¿œï¿œo, and A. P. Moreira, "Nonlinear model predictive formation control: An iterative weighted tuning approach," *J. Intell. Robot. Syst.*, vol. 80, no. 3-4, pp. 441–454, 2015.

[16] A. Mondal, C. Bhowmick, L. Behera, and M. Jamshidi, "Trajectory tracking by multiple agents in formation with collision avoidance and connectivity assurance," *IEEE Syst. J.*, vol. 12, no. 3, pp. 2449–2460, 2018.

[17] H.-S. Shin, A. Garcia, and S. Alvarez, "Information-driven persistent sensing of a non-cooperative mobile target using UAVs," *J. Intell. Robot. Syst.*, vol. 92, 12 2018.

[18] D. Pickem, P. Glotfelter, L. Wang, M. Mote, A. Ames, E. Feron, and M. Egerstedt, "The robotarium: A remotely accessible swarm robotics research testbed," in *2017 IEEE Int. Conf. Robot. Autom.*, 2017, pp. 1699–1706.

[19] A. D. Ames, S. Coogan, M. Egerstedt, G. Notomista, K. Sreenath, and P. Tabuada, "Control barrier functions: Theory and applications," in *2019 Eur. Control Conf.*, 2019, pp. 3420–3431.

[20] C. Tomlin, G. Pappas, and S. Sastry, "Conflict resolution for air traffic management: A study in multiagent hybrid systems," *IEEE Trans. on Autom. Control*, vol. 43, no. 4, pp. 509–521, 1998.

[21] A. D. Ames, J. W. Grizzle, and P. Tabuada, "Control barrier function based quadratic programs with application to adaptive cruise control," in *IEEE Conf. Decis. Control*, 2014, pp. 6271–6278.

[22] Q. Nguyen and K. Sreenath, "Safety-critical control for dynamical bipedal walking with precise footstep placement," *IFAC-PapersOnLine*, vol. 48, no. 27, pp. 147 – 154, 2015, analysis and Design of Hybrid Systems ADHS.

[23] T. Gurriet, A. Singletary, J. Reher, L. Ciarletta, E. Feron, and A. Ames, "Towards a framework for realizable safety critical control through active set invariance," in *2018 ACM/IEEE Int. Conf. Cyber-Phys. Syst.*, 2018, pp. 98–106.

[24] A. Li, L. Wang, P. Pierpaoli, and M. Egerstedt, "Formally correct composition of coordinated behaviors using control barrier certificates," in *2018 IEEE/RSJ Int. Conf. Intell. Robot. Syst.*, 2018, pp. 3723–3729.

[25] W. Luo and K. Sycara, "Voronoi-based coverage control with connectivity maintenance for robotic sensor networks," in *2019 Int. Symp. Multi-Robot Multi-Agent Syst.*, 2019, pp. 148–154.

[26] L. Wang, A. D. Ames, and M. Egerstedt, "Multi-objective compositions for collision-free connectivity maintenance in teams of mobile robots," in *2016 IEEE Conf. Decis. Control*, 2016, pp. 2659–2664.

[27] D. Han and D. Panagou, "Robust multitask formation control via parametric lyapunov-like barrier functions," *IEEE Trans. Autom. Control*, vol. 64, no. 11, pp. 4439–4453, 2019.

[28] U. Borrmann, L. Wang, A. D. Ames, and M. Egerstedt, "Control barrier certificates for safe swarm behavior," *IFAC-PapersOnLine*, vol. 48, no. 27, pp. 68 – 73, 2015, analysis and Design of Hybrid Systems ADHS.

[29] L. Wang, A. Ames, and M. Egerstedt, "Safety barrier certificates for heterogeneous multi-robot systems," in *2016 Am. Control Conf.*, 2016, pp. 5213–5218.

[30] E. A. Wan and R. van der Menve, "The unscented kalman filter for nonlinear estimation," in *Proc. IEEE 2000 Adaptive Syst. Signal Process., Commun., Control Symp.*, Lake Louise, Alberta, Canada, Oct. 2000, pp. 153–158.

[31] X. Xu, P. Tabuada, J. W. Grizzle, and A. D. Ames, "Robustness of control barrier functions for safety critical control," *IFAC-PapersOnLine*, vol. 48, no. 27, pp. 54 – 61, 2015, analysis and Design of Hybrid Systems ADHS.

[32] M. Endo, T. Ibuki, and M. Sampei, "Collision-free formation control

for quadrotor networks based on distributed quadratic programs," in *2019 Am. Control Conf.*, 2019, pp. 3335–3340.